%% file: main.tex
\documentclass[10pt,twocolumn,letterpaper]{article}

\usepackage[pagenumbers]{cvpr} 

\usepackage{blindtext}
\usepackage{graphicx}
\usepackage{caption}

\input{preamble}
\definecolor{cvprblue}{rgb}{0.21,0.49,0.74}
\usepackage[pagebackref,breaklinks,colorlinks,allcolors=cvprblue]{hyperref}

\usepackage{amsmath}
\usepackage{amsfonts}
\usepackage{booktabs}              
\usepackage{tabularx}              
\usepackage{multirow}              

\newcommand{\thoughtlogo}{\raisebox{-.3ex}{\includegraphics[height=1.8ex]{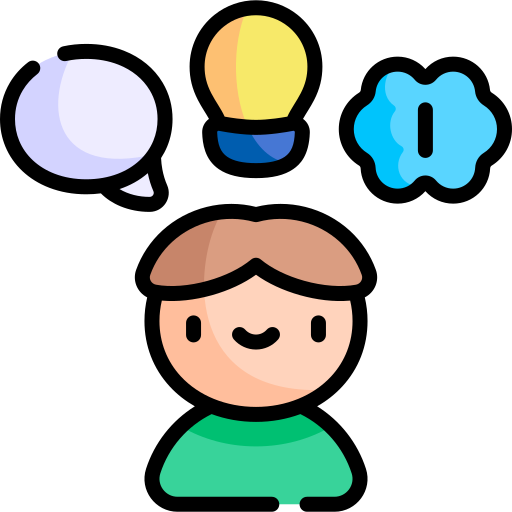}}}

\def\paperID{*****} 
\def\confName{CVPR}
\def\confYear{2026}

\title{{\scalebox{2}{\thoughtlogo}} \textit{Training Multimodal Large Reasoning Models Needs Better Thoughts}:\\
A Three-Stage Framework for Long Chain-of-Thought Synthesis and Selection}

\author{Yizhi Wang$^{1, 2}$, Linan Yue$^{1, 2, *}$, Min-Ling Zhang$^{1, 2, *}$\\
School of Computer Science and Engineering, Southeast University$^1$\\
Key Laboratory of Computer Network and Information Integration (SEU), Ministry of Education$^2$\\
Nanjing {\rm 210096}, China\\
{\tt\small wang\_yz@seu.edu.cn, lnyue@seu.edu.cn, zhangml@seu.edu.cn}
}

\begin{document}
\twocolumn[{
\renewcommand\twocolumn[1][]{#1}%
\maketitle
\begin{center}
\centering
\includegraphics[width=0.8\textwidth]{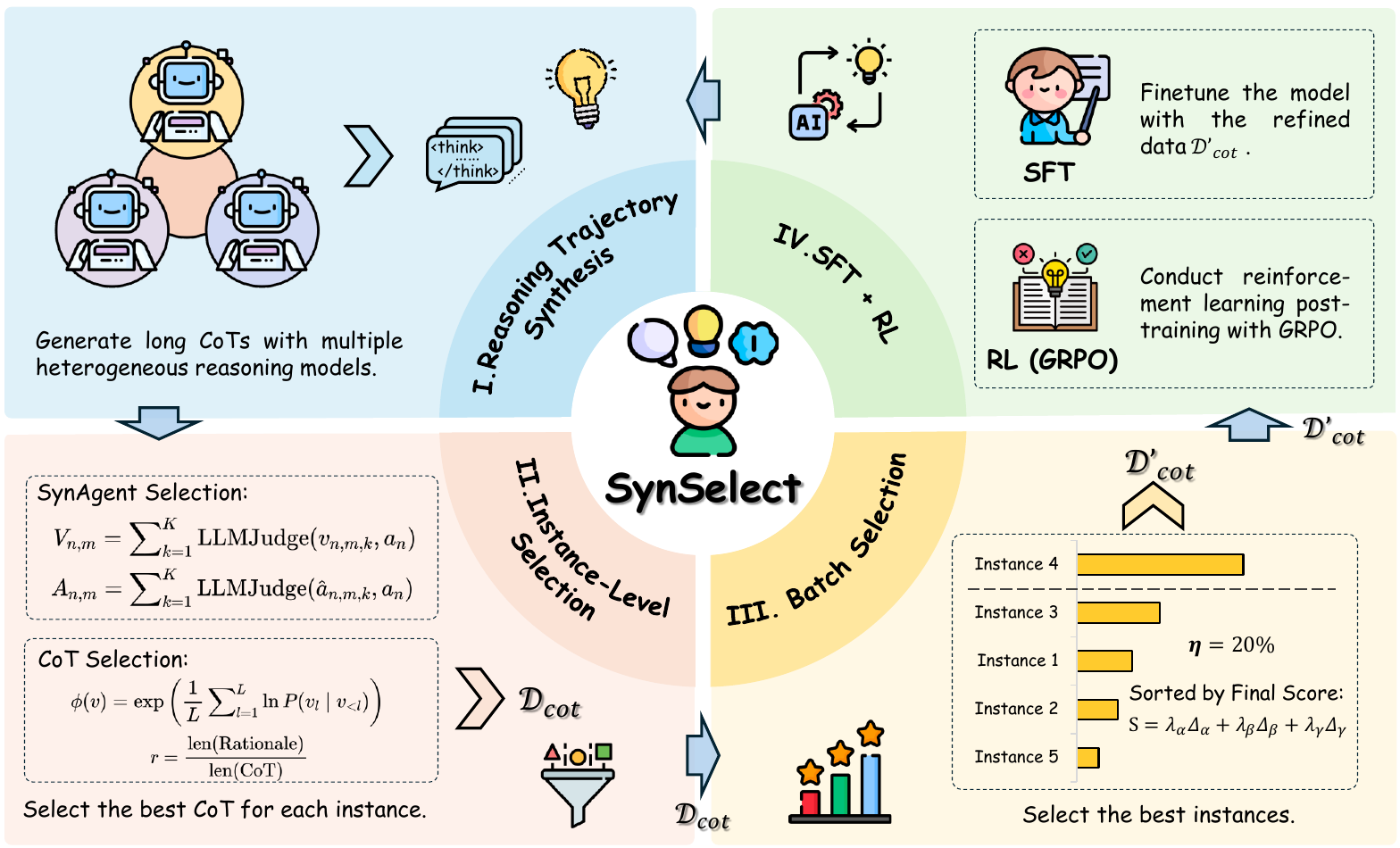}
\captionof{figure}{Overview of the \textit{SynSelect} pipeline. \textit{SynSelect} first generates high-quality multimodal CoT data in three stages: (I) Reasoning trajectory synthesis: multiple heterogeneous MLRMs generate candidate CoTs; (II) Instance-level selection: the best CoT is picked for each question; (III) Batch selection: a optimal subset is selected from all instances. 
Then, in (IV), the synthesized dataset is used for SFT, followed by RL post-training to further enhance multimodal reasoning capabilities.
Finally, the trained model can subsequently be reused as a new MLRM in Stage (I), enabling iterative data synthesis while keeping the training procedure decoupled from \textit{SynSelect} .}
\end{center}
}]

\input{sec/0_abstract}
\input{sec/1_intro}
\input{sec/2_related_works}

\input{sec/3_problem_definition} 
\input{sec/4_method}

\input{sec/5_experiments}
\input{sec/6_conclusion}

{
    \small
    \bibliographystyle{ieeenat_fullname}
    \bibliography{main}
}

\setcounter{section}{0}
\setcounter{subsection}{0}
\setcounter{figure}{0}
\setcounter{equation}{0}
\setcounter{table}{0}
\input{sec/X_supp}

\end{document}

%% file: sec/0_abstract.tex
\begin{abstract}
Large Reasoning Models (LRMs) have demonstrated remarkable performance on complex reasoning tasks through long Chain-of-Thought (CoT) reasoning. Extending these successes to multimodal reasoning remains challenging due to the increased complexity of integrating diverse input
modalities and the scarcity of high-quality long CoT training data. Existing multimodal datasets and CoT synthesis 
\noindent\rule{0.15\textwidth}{0.4pt}\\
\noindent\footnotesize\textsuperscript{*}Corresponding authors.\\
\normalsize
methods still suffer from limited reasoning depth, modality conversion errors, and rigid generation pipelines, hindering model performance and stability. To this end, in this
paper, we propose \textit{SynSelect}, a novel three-stage \textit{Syn}thesis-\textit{Select}ion framework for generating high-quality long CoT data tailored to multimodal reasoning tasks. Specifically, \textit{SynSelect} first leverages multiple heterogeneous multimodal LRMs to produce diverse candidate CoTs, and then applies both instance and batch level selection to filter high-quality CoTs that can effectively enhance the model’s reasoning capabilities.
Extensive experiments on multiple multimodal benchmarks demonstrate that models supervised fine-tuned on \textit{SynSelect}-generated data significantly outperform baselines and achieve further improvements after reinforcement learning post-training. Our results validate \textit{SynSelect} as an effective approach for advancing multimodal LRMs reasoning capabilities. 
Code and synthesized dataset are available at \url{https://github.com/Snowstorm1492/SynSelect}.
\end{abstract}

%% file: sec/1_intro.tex
\section{Introduction}
\label{sec:intro}

\begin{figure}
    \centering
    \includegraphics[width=0.9\linewidth]{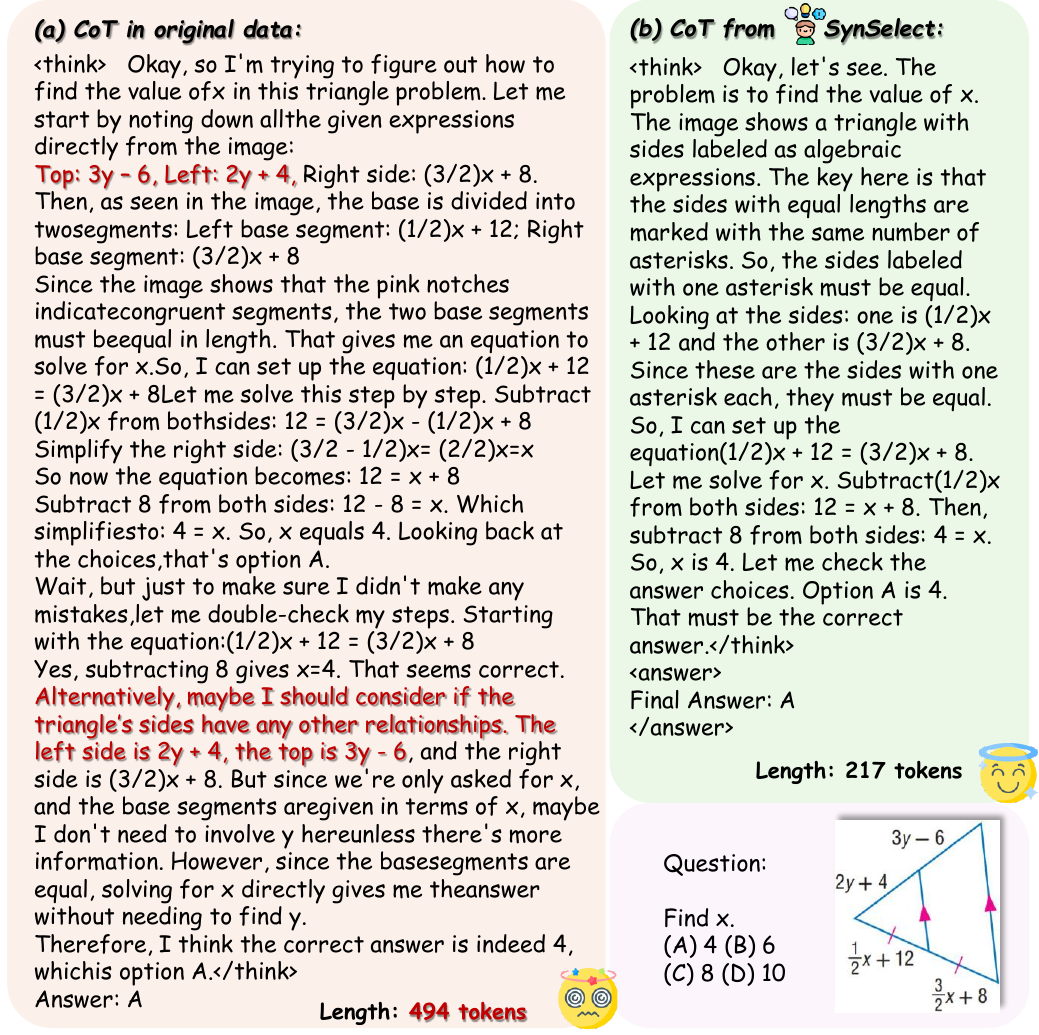}
    \caption{Example of \textit{SynSelect} data. SynSelect produces shorter and more accurate CoTs compared with the original data, reducing redundancy and errors through its synthesis–selection framework. More examples could be found in appendix.}
    \label{fig:example}
\end{figure}

Large Reasoning Models (LRMs), such as DeepSeek-R1 \cite{deepseek-ai_deepseek-r1_2025}, have exhibited exceptional capabilities in complex reasoning tasks, attributed to the application of large-scale reinforcement learning (RL) post-training in language models \cite{openai_openai_2024, deepseek-ai_deepseek-v3_2025, xie_logic-rl_2025, qwq-32b-preview, deepseek-ai_deepseek-r1_2025, yang2025qwen3}. These models typically employ structured long Chain-of-Thought (CoT) reasoning, enhanced by mechanisms of self-reflection and self-correction, which significantly improve their logical reasoning and decision-making abilities. Inspired by this success, researchers have begun exploring the extension of LRMs training paradigms to the multimodal domain.
However, existing studies reveal that current multimodal LRMs (MLRMs) still face challenges in terms of both performance and stability \cite{zhang2024mm, meng2025mm, yin2024survey, yang2025r1, li_system_2025, huang2025vision}. This is primarily due to the increased architectural complexity and training difficulty associated with integrating diverse modalities, such as language and vision. Under such conditions, RL post-training often fails to converge stably in multimodal settings and may even lead to performance degradation, ultimately hindering the model’s effectiveness in complex reasoning tasks.

To mitigate these challenges, a promising strategy is to first employ supervised fine-tuning (SFT) methods on long CoTs to improve the model’s basic reasoning ability, followed by RL-based optimization to further boost performance \cite{huang2025vision, yang2025r1, chen2025advancing, chen2025g1, peng2025lmm}. This two-stage paradigm has proven effective in text-only settings.
However, extending it to multimodal reasoning introduces new obstacles. Specifically, multimodal reasoning demands precise alignment between visual content and linguistic logic at every step, making model performance highly sensitive to the quality of SFT data. Unfortunately, existing multimodal datasets generally lack high-quality, long CoT annotations. For example, as shown in Figure~\ref{fig:example} (a), they could contain CoTs that are verbose or invalid. Current benchmarks (e.g., the multimodal subset of MMLU \cite{hendrycks2020measuring}, MMBench \cite{liu2024mmbench}, ScienceQA \cite{lu2022learn}) focus on basic tasks and offer limited support for complex reasoning chains. Furthermore, prevailing CoT synthesis approaches also encounter several limitations. Some methods first convert visual inputs into textual descriptions and then apply models like DeepSeek-R1 for CoT generation, which may introduce cascading errors due to modality conversion \cite{huang2025vision,yang2025r1}. Others, such as LLaVA-CoT \cite{xu_llava-cot_2025}, generate CoTs in fixed stages, often resulting in rigid and less flexible reasoning processes.

To this end, in this paper, we propose \textbf{\textit{SynSelect}}, a three-stage \textbf{\textit{Syn}}thesis-\textbf{\textit{Select}}ion framework designed to generate high-quality long CoT data tailored for multimodal reasoning tasks. The framework contains three step-wise stages: Synthesis, Initial Filtering, and Refined Selection, aiming to construct training data for the SFT of MLRMs. Specifically,
\textbf{\textit{in the Synthesis stage}}, given a dataset of $N$ instances, {\textit{SynSelect}} leverages $M$ heterogeneous MLRMs to perform $K$ independent samplings per instance, yielding $M \times K$ diverse reasoning paths. By exploiting architectural diversity and sampling variance, {\textit{SynSelect}} significantly enhances the coverage and diversity of candidate CoTs.
\textbf{\textit{During the Initial Filtering stage}}, {\textit{SynSelect}} applies multi-dimensional evaluation metrics in the instance-level selection, including answer correctness, reasoning validity, and length appropriateness, to assess the quality of each candidate CoT. It then selects the most semantically clear and logically coherent reasoning path for each instance.
\textbf{\textit{In the Refined Selection stage}}, {\textit{SynSelect}} adopts a batch selection strategy, introducing query-aware and CoT-aware selection  to identify the most instructive $N'$ samples from the original $N$ instances. These samples are optimized for boosting reasoning capabilities, achieving a balance between training efficiency and model performance. The final output is a high quality dataset specifically curated for complex multimodal reasoning. As shown in Figure~\ref{fig:example} (b), \textit{SynSelect} ultimately produces concise, accurate, and well-structured CoTs.

Our contributions are summarized as follows:
\begin{itemize}
\item We propose \textit{SynSelect}, a three-stage synthesis-selection framework to generate high-quality long CoT data tailored for multimodal reasoning tasks.
\item In the synthesis stage, we employ multiple heterogeneous MLRMs to generate long CoTs, effectively leveraging architectural diversity to produce more comprehensive and diverse reasoning paths.
\item We design multi-level selection strategies at both the instance and batch levels, enabling fine-grained control over CoT quality. The selected samples can contribute to improving reasoning capabilities.
\item Extensive experiments are conducted on multiple multimodal benchmark datasets. Results show that models supervised fine-tuned on \textit{SynSelect}-generated long CoT data outperform various baselines on complex reasoning tasks. Besides, models fine-tuned with \textit{SynSelect} data achieve additional gains after RL post-training compared to models trained solely with RL. These findings validate the effectiveness of \textit{SynSelect}-generated data.
\end{itemize}

%% file: sec/2_related_works.tex
\section{Related Works}
\label{sec:related_works}

\textbf{Multimodal Large Reasoning Models.}
In recent years, Multimodal Large Language Models (MLLMs) have attracted considerable attention and achieved promising results across a range of tasks \cite{bai_qwen25-vl_2025, dubey_llama_2024, hurst2024gpt, li_llava-onevision_2024, Qwen2VL, thawakar_llamav-o1_2025, wu_deepseek-vl2_2024, xu_llava-cot_2025, zhang_mm15_2024, yao_minicpm-v_2024, chen2024expanding}. However, these models often lack explicit long Chain-of-Thought (CoT) reasoning mechanisms \cite{bai_qwen25-vl_2025, dubey_llama_2024, Qwen2VL, wu_deepseek-vl2_2024, chen2024expanding}, which poses significant challenges when tackling complex tasks that require multi-step reasoning.
Inspired by the continual progress of LLMs in reasoning (e.g., DeepSeek-R1), which enhances reasoning strategies via large-scale RL post-training, researchers have begun to focus on developing MLRMs \cite{team_kimi-vl_2025, qvq-72b-preview, step3blog, wang2025skywork, deng2025openvlthinker, shen2025vlm, zhang_r1-vl_2025, wang_enhancing_2025, yao_mulberry_2024}.
Current reasoning models can be broadly categorized into two types:
(1) RL-based training approaches \cite{zhou2025r1, meng2025mm, wang2025skywork, zhang_r1-vl_2025, peng2025lmm, chen2025g1}. These methods (e.g., MM-Eureka \cite{meng2025mm} and VisualThinker R1 \cite{zhou2025r1}) directly employ RL methods to train models for better reasoning capabilities. While promising, they often suffer from issues such as shallow reasoning trajectories \cite{huang2025vision, meng2025mm}, visual hallucinations \cite{wang2025skywork}, and ineffective length-based rewards \cite{zhou2025r1}, leading to suboptimal reasoning quality and generalization.
(2) SFT+RL approaches \cite{chen2025advancing, yang2025r1, huang2025vision, tan2025reason}. These methods (e.g., R1-OneVision \cite{yang2025r1} and Vision-R1 \cite{huang2025vision}) typically rely on converting visual inputs into intermediate textual representations, such as natural language or code, which are then fed into LRMs to generate long CoTs for SFT. RL is then applied on top of these fine-tuned models. However, such conversion pipelines are prone to losing or distorting crucial visual information, which may degrade the accuracy and consistency of the synthesized long CoTs, ultimately impairing the model’s performance.

\noindent
\textbf{Multimodal Reasoning Datasets Construction.}
Constructing high-quality multimodal reasoning datasets is crucial for enabling complex reasoning capabilities in models. Early datasets (e.g., ChartQA \cite{masry2022chartqa}, MMBench \cite{liu2024mmbench}, and ScienceQA \cite{lu2022learn}) primarily focus on foundational tasks like visual question answering and chart understanding. While these datasets exhibit a certain degree of complexity, they generally lack explicit long CoT reasoning processes, making it difficult for models to perform multi-step reasoning.
To address this limitation, researchers have explored automatic methods \cite{wang2024enhancing,yao2024mulberry,xu_llava-cot_2025} for constructing multimodal reasoning data. For example, LLava-CoT-100K \cite{xu_llava-cot_2025} leverages staged prompting with GPT-4o to generate reasoning CoT composed of four structured components: Summary, Caption, Reasoning, and Conclusion. This improves the alignment between visual content and language-based reasoning. However, its rigid, predefined reasoning structure limits the model's ability to generate flexible and diverse reasoning paths.
More recent approaches introduce text-based reasoning models to improve the quality of multimodal CoT generation \cite{chen2025sft,yang2025r1,huang2025vision}. For example, R1-OneVision-Dataset \cite{yang2025r1} first translates images into formal representations (e.g., matplotlib code), which are then processed by LRMs to generate CoTs. Vision-R1-Cold \cite{huang2025vision} employs modality bridging techniques to convert visual inputs into natural language, indirectly guiding the reasoning process. While such methods improve the quality of synthesized CoTs, they still face challenges: the vision-to-text conversion may lose critical details or introduce semantic errors, leading to inconsistencies between the CoT and the original visual content, which undermines the accuracy and stability of the reasoning chain.
To overcome these limitations, in this paper, we propose \textit{SynSelect}, a long CoT synthesis-selection framework tailored for multimodal reasoning tasks.

%% file: sec/3_problem_definition.tex
\section{Problem Definition}

In this section, we formally define the problem addressed in this paper. Specifically, given a raw multimodal dataset $\mathcal{D}_\text{raw} = \{(q_n, a_n) | 1 \leq n \leq N, n \in \mathbb{N}_{+}\}$, where each input query $q_n$ consists of a pair $(\text{query}, \text{image})$, and $a_n$ denotes the corresponding answer, our objective is two-fold:
\begin{itemize}
\item \textbf{Reasoning Data Construction:} Augment each raw sample with a long CoT reasoning path with ``\textit{aha memont}'', resulting in a new dataset:
\begin{equation}
\mathcal{D}_\text{cot} = \{(q_n, a_n, \mathrm{CoT}_n) | 1 \leq n \leq N, n \in \mathbb{N}_{+}\}.
\end{equation}
\item \textbf{Subset Refinement:} Select a high-quality subset of $\mathcal{D}_\text{cot}$ with size $N' \ll N$, leading to a compact but informative reasoning dataset:
\begin{equation}
    \mathcal{D}'_\text{cot} = \{(q_n, a_n, \mathrm{CoT}_n) | 1 \leq n \leq N', n \in \mathbb{N}_{+}\},
\end{equation}
where we define the selection ratio as $\eta = N' / N$.
\end{itemize}

Based on the constructed dataset, we perform SFT on MLLMs to enhance their reasoning capabilities. Besides, RL methods (e.g., GRPO \cite{shao2024deepseekmath}) can be applied to the SFT-tuned models, ultimately yielding MLRMs with stronger step-by-step reasoning abilities.

\begin{figure}[t]
    \centering
    \includegraphics[width=7.3cm]{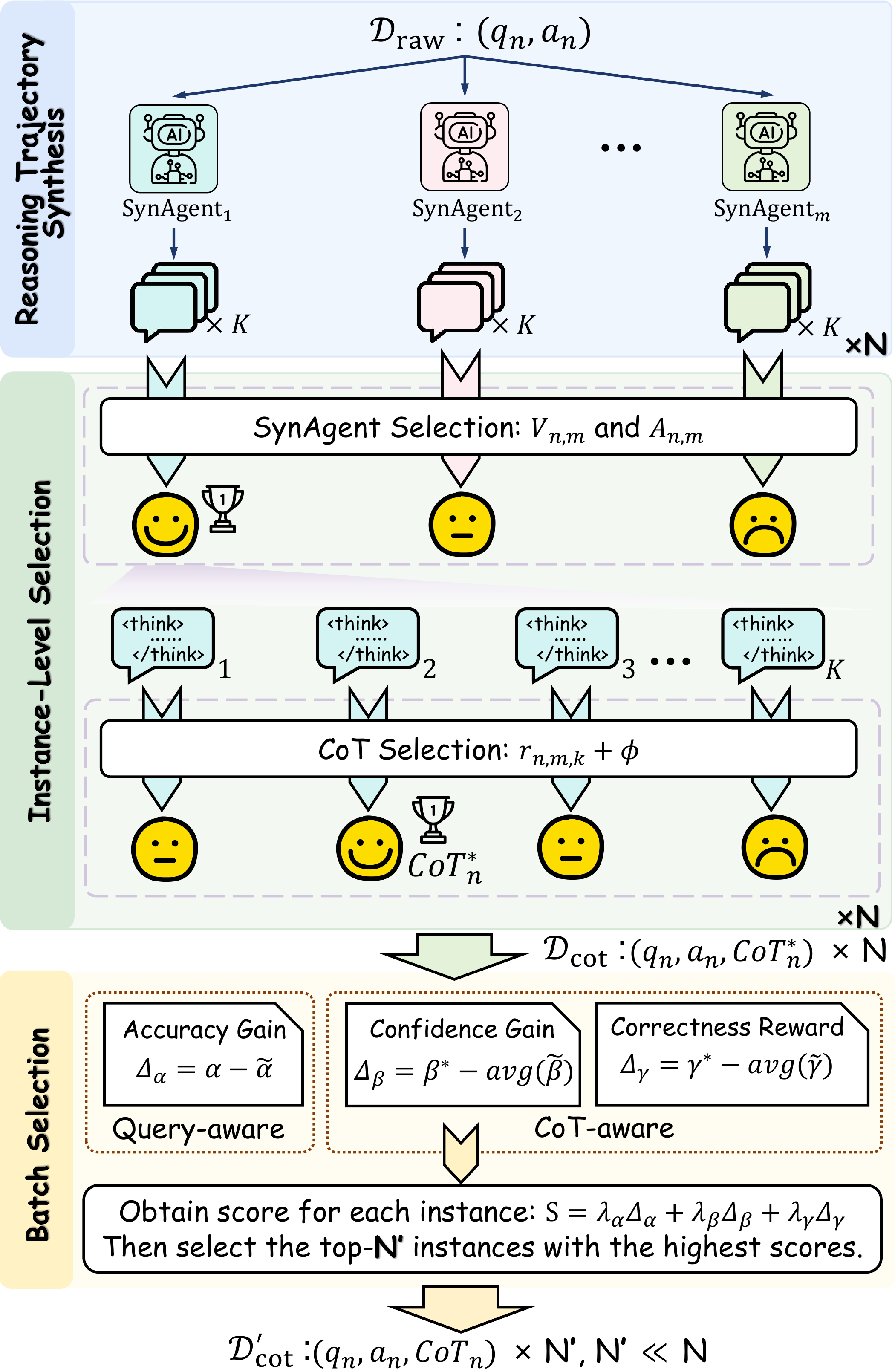}
    \caption{Architecture of the proposed \textit{SynSelect} framework, including Reasoning Trajectory Synthesis, Instance-level Selection, and Batch Selection. $avg(\cdot)$ denotes the average computed over both $M$ and $K$. Detailed calculation is in Section~\ref{sec:cot-aware}.}
    \label{fig:framework}
\end{figure}

%% file: sec/4_method.tex
\section{Methodology}

In this section, as shown in Figure \ref{fig:framework}, we present the details of our proposed \textbf{\textit{SynSelect}} framework, which consists of three sequential stages:

\noindent
\textbf{1. Synthesis Stage: Reasoning Trajectory Synthesis.}
For each instance in the raw multimodal dataset $\mathcal{D}_\text{raw}$, we first generate a set of diverse long CoT reasoning paths.

\noindent
\textbf{2. Initial Filtering Stage: Instance-level Selection.}
We then select the most appropriate reasoning path for each instance from its all CoT candidates. This stage takes both the generated CoTs and the original dataset $\mathcal{D}_\text{raw}$ as input and yields the full multimodal reasoning dataset $\mathcal{D}_\text{cot}$.

\noindent
\textbf{3. Refined Selection Stage: Batch Selection.}
We perform selection at the dataset level. Specifically, we treat the entire dataset as a batch and take $\mathcal{D}_\text{cot}$ as input to identify a refined subset $\mathcal{D}'_\text{cot} \subset \mathcal{D}_\text{cot}$, consisting of high-quality reasoning samples that are most beneficial for training MLRMs.

\subsection{Reasoning Trajectory Synthesis}
To generate long CoTs for each instance, a straightforward approach  is to directly leverage existing MLRMs for generation. However, current MLRMs exhibit diverse capability profiles, with varying strengths across different tasks. To avoid over-reliance on a single model’s behavior, we employ a set of $M$ heterogeneous MLRMs as \textit{synthesis agents}, denoted as $\text{SynAgent}_m(\cdot)$, to collaboratively generate CoTs. 

Specifically, for each $q_n$, each $\text{SynAgent}_m(\cdot)$ performs $K$ rounds of inference with different random seeds $\varepsilon_{n,m,k}$:
\begin{equation}
(\text{CoT}, \hat{a})_{n, m, k} = \text{SynAgent}_m(q_n; \varepsilon_{n,m,k}),
\end{equation}
where $\hat{a}$ is the model’s predicted answer, and $\text{CoT}$ denotes the corresponding reasoning path. Performing $K$ rounds of sampling allows us to better explore the model's reasoning space,  yielding diverse reasoning trajectories.
In this way, we can obtain $M \times K$ reasoning paths per instance, resulting in a total of $N \times M \times K$ CoTs across the entire dataset. 

\noindent
\textbf{\textit{Remark}.}
All \textit{synthesis agents} used in this stage are native MLRMs (e.g., R1-OneVision \cite{yang2025r1}, MM-Eureka \cite{meng2025mm} and Vision-R1 \cite{huang2025vision}). Compared to template-based or vision-text concatenation approaches, these models exhibit more flexible reasoning processes and emergent behaviors such as self-reflection. Furthermore, their native multimodal support enables more effective modality fusion and avoids the risk of information loss that may arise during intermediate modality conversion (e.g., image-to-caption).

\subsection{Instance-level Selection}
\label{sec:instance-level_select}

Although the synthesized candidate pool contains a substantial number of high-quality CoTs, the majority remain ordinary or suboptimal. Therefore, it is essential to identify the most appropriate reasoning path for each instance.
However, evaluating the quality of CoTs is inherently challenging. In our setting, the primary objective is to ensure that the selected CoTs contribute effectively to the SFT of MLLMs. To this end, we design three key indicators to guide the selection process:
(1) \textit{Answer correctness},
(2) \textit{Reasoning validity}, and
(3) \textit{Length appropriateness}.

Based on these indicators, a straightforward selection strategy for each query is to aggregate all $M \times K$ CoTs and directly choose the best one. However, our preliminary experiments revealed that CoTs generated by the same \textit{synthesis agent} tend to exhibit higher semantic similarity. For a given query, identifying the agent that performs best often leads to more reliable CoT selection. 
Therefore, we adopt a hierarchical instance-level selection strategy. Specifically, for each instance:
\begin{itemize}
    \item SynAgent Selection: We first identify the best-performing agent $\text{SynAgent}_{m^{*}}$ based on the defined indicators.
    \item CoT Selection: Then, we select the optimal reasoning path $k^{*}$ from the $K$ CoTs generated by agent $\text{SynAgent}_{m^{*}}$.
\end{itemize}
Below, we introduce the designed indicators, followed by details of our hierarchical instance-level selection strategy.

\subsubsection{Key Indicators of Instance-level Selection} In this subsection, we define three key indicators:
\paragraph{\textit{(1) Answer Correctness}} Answer correctness is a foundational guarantee for CoT quality. In this indicator, we use a LLM Judge to verify whether the predicted answer $\hat{a}$ is consistent with the ground-truth answer~$a$:
\begin{equation}
    \text{LLMJudge}(\hat{a}, a) = 
    \begin{cases} 
    1 & \text{if } \hat{a} \text{ is consistent with } a, \\
    0 & \text{others}.
    \end{cases}.
    \label{LLMJudge}
\end{equation}
If the answer associated with a CoT is incorrect (i.e., $\text{LLMJudge}(\hat{a}, a)=0$), the CoT is immediately discarded. Otherwise, it is retained for further evaluation.

\paragraph{\textit{(2) Reasoning Validity}} 
Our preliminary experiments revealed that several CoTs yield correct final answers but contain problems such as flawed causal logic and inaccurate visual grounding. Incorporating such CoTs into the SFT training set may degrade model performance, making it essential to identify and remove them.
To address this, we propose a novel method for verifying reasoning validity. Specifically, we leverage a small-scale, non-reasoning model (e.g., Qwen2.5-VL-3B \cite{bai_qwen25-vl_2025}), referred to as the LLMPlayer, and examine whether it can correctly solve the problem when aided by a specific CoT generated by the synthesis agents. The intuition is that a reliable CoT should provide effective guidance and improve the performance of a weaker model.

It is worth noting that this step aims to perform a coarse-grained filtering of \textit{potentially unreliable CoTs}. We adopt the  assumption: If the LLMPlayer fails to produce the correct answer when provided with a CoT, the CoT is likely to be unreliable. Conversely, if the LLMPlayer succeeds with the CoT, even if it could answer correctly without it, we tentatively consider the CoT to be reliable. A more fine-grained selection process will be conducted in subsequent stages.

This verification process allows us to indirectly evaluate the quality of CoTs through their influence on model behavior, without explicitly dissecting their internal logic. Formally, we define the output and confidence score of the LLMPlayer, which are further utilized in the hierarchical instance-level selection strategy:
\begin{equation}
{v}_{n, m, k} = \text{LLMPlayer}(q_n, \mathrm{CoT}_{n,m,k};\varepsilon_{n,m,k}),
\end{equation}
where $v_{n,m,k}$ is the output answer generated by the LLMPlayer.
The corresponding confidence score is defined as:
\begin{equation}
\phi(v) = \exp{\left(\frac{1}{L} \sum\nolimits_{l=1}^{L} \ln P(v_l \mid v_{<l})\right)},
\label{confidence}
\end{equation}
where $\phi(v)$ represents the sequence-level confidence of the generated answer $v$ (e.g. $v_{n,m,k}$), computed as the exponential of the average log-likelihood over the sequence length~$L$.


\paragraph{\textit{(3) Length Appropriateness}} 
 In practice, we observe that existing MLRMs tend to generate overly verbose CoTs for certain questions. These CoTs often include not only useful reasoning steps but also redundant self-verifications and meaningless repetitions. Given the limited reasoning capabilities of current multimodal models, assessing the redundancy of a CoT becomes particularly important.

However, although some CoTs are excessively long, complex questions inherently require more elaborate reasoning, rendering simple length-based filtering ineffective. To strike a balance, we propose a discrimination method based on the rationale ratio. Specifically, we employ a LLM-based Extractor to identify the core reasoning steps (i.e., the Rationale) within each CoT, and then compute the ratio between the rationale length and the total CoT length:
\begin{equation}
r = \frac{\text{len}(\text{Rationale})}{\text{len}(\text{CoT})},
\end{equation}
where a higher ratio $r$ indicates that the CoT contains denser reasoning content and less redundancy.

\subsubsection{Hierarchical Instance-level Selection Strategy}
After defining the above indicators, we proceed to select the high-quality CoT for each instance.
As previously mentioned, in SynAgent Selection, we begin by selecting the best-performing agent $\text{SynAgent }_{m^{*}}$.

First, we define the total number of correct responses provided by the $m$-th synthesis agent (which generates $K$ CoTs for the $n$-th sample) as:
\begin{equation}
A_{n,m} = \sum\nolimits_{k=1}^{K} \text{LLMJudge}(\hat{a}_{n,m,k}, a_n).
\end{equation}
This metric reflects the agent’s ability to produce correct answers, i.e., among the $K$ responses, how many match the reference answer. This corresponds to \textit{Answer Correctness}.

Next, we define the total number of valid CoTs synthesized by the same agent as:
\begin{equation}
V_{n,m} = \sum\nolimits_{k=1}^{K} \text{LLMJudge}(v_{n,m,k}, a_n).
\end{equation}
This metric reflects the reliability of the reasoning paths, which corresponds to \textit{Reasoning Validity}.

Finally, we rank the $m$ agents based on their $V_{n,m}$ scores. If two or more agents exhibit comparable reliability (i.e., similar $V_{n,m}$), we break the tie by further ranking them based on their answer correctness $A_{n,m}$.

After determining the \textit{synthesis agent}, in CoT Selection, we first discard~all CoTs generated by the selected agent with incorrect answers (i.e., $\text{LLMJudge}(\hat{a}, a)=0$).
Next, we employ both \textit{Reasoning Validity} and \textit{Length Appropriateness} to identify the highest-scoring CoT based on:
\begin{equation}
    k^* = \arg\max_k \left(\phi({v}_{n,\ m^*,k})  + \lambda_{k} \cdot r_{n,\ m^*,k}\right),
\end{equation}
where $\phi({v}_{n,\ m^*,k})$ is the confidence of the small model's responses and $r_{n,\ m^*,k}$ measures the length appropriateness.

As a result, for each instance $q_n$, we can obtain its corresponding optimal reasoning trajectory $\mathrm{CoT}_{n,\ m^{*},k^{*}}$, thereby constructing the multimodal reasoning dataset $\mathcal{D}_\text{cot}$.

\subsection{Batch Selection}

Appropriate SFT data can play a crucial role in the early stages of MLRMs training. It helps guide the model toward learning correct reasoning patterns and facilitates more significant performance gains after RL post-training. However, this guiding effect depends more on quality than quantity. Therefore, at this stage,
given the full dataset $\mathcal{D}_\text{cot}$, we aim to select the most instructive $N'$ samples out of the total $N$ instances, which are most helpful for fostering reasoning capabilities in the model.
Since each instance in our generated dataset consists of a query and a corresponding CoT, we perform batch selection from both the query perspective and the CoT perspective with the LLMPlayer in \textit{Reasoning Validity}.

\begin{table*}[!ht]
\centering

\scriptsize

\begin{tabularx}{\linewidth}{
  >{\raggedright\arraybackslash}X  
  >{\centering\arraybackslash}X    
  >{\centering\arraybackslash}X    
  >{\centering\arraybackslash}X    
  >{\centering\arraybackslash}X    
}
\toprule
\textbf{Model} & \textbf{MathVerse{\scriptsize-ALL}} & \textbf{MathVerse{\scriptsize-Vision Only}} & \textbf{MathVista} & \textbf{WeMath} \\
\midrule
\multicolumn{5}{l}{\textit{\textbf{Base Model}}} \\  
Qwen2.5-VL-7B  & 43.6 & 38.2 & 63.7 & 61.0 \\
\midrule
\multicolumn{5}{l}{\textit{\textbf{Other Models}}} \\  
GPT-4o  & 41.2 & 34.5 & 60.0 & 69.0 \\
InternVL2.5-8B  & 35.6 & 22.8 & 64.5 & 53.8 \\
Qwen2-VL-7B  & 30.2 & 25.4 & 61.6 & 50.5 \\
LLaVA-CoT-11B  & - & 22.6 & 52.5 & - \\
\midrule
\multicolumn{5}{l}{\textit{\textbf{SFT Only}}} \\  
R1-Onevision(rpt) & 43.4 & 39.7 & - & - \\
R1-Onevision & 41.9$\pm$0.6 & 38.0$\pm$0.9 & 59.9$\pm$2.3 & 57.6$\pm$0.6 \\

\textit{SynSelect}-$\mathcal{D}_\text{cot}$ & 44.2$\pm$0.7 & 40.2$\pm$0.7 & 63.1$\pm$1.9 & 60.2$\pm$1.3 \\
\textit{SynSelect}-$\mathcal{D}'_\text{cot}$ & \textbf{45.2$\pm$0.7} & \textbf{41.7$\pm$1.0} & \textbf{64.3$\pm$0.8} & \textbf{61.8$\pm$0.6} \\
\midrule
\multicolumn{5}{l}{\textit{\textbf{RL Only}}} \\  
Qwen2.5-VL-7B & 43.9$\pm$0.8 & 39.2$\pm$0.7 & 62.8$\pm$1.4 & 59.1$\pm$1.1 \\
\midrule
\multicolumn{5}{l}{\textit{\textbf{SFT + RL}}} \\  
R1-Onevision(rpt) & \textbf{46.4} & 40.0 & 64.1 & 61.8 \\
R1-Onevision & 44.6$\pm$0.3 & 40.0$\pm$0.8 & 64.2$\pm$0.6 & 59.0$\pm$0.7 \\

\textit{SynSelect}-$\mathcal{D}_\text{cot}$ & 45.0$\pm$0.4 & 40.3$\pm$0.5 & \textbf{66.0$\pm$1.4} & 61.2$\pm$0.9 \\
\textit{SynSelect}-$\mathcal{D}'_\text{cot}$ & 46.1$\pm$0.3 & \textbf{41.9$\pm$1.1} & 65.7$\pm$1.2 & \textbf{62.8$\pm$1.4} \\
\bottomrule
\end{tabularx}
\caption{Experimental results on multimodal reasoning benchmarks. Except for models trained on the R1-Onevision-Dataset and \textit{SynSelect} data, all results are taken from the original R1-Onevision paper, with ``R1-Onevision (rpt)'' indicates the reported values. Reproduced experiments are reported as (mean$\pm$std). The highest accuracy in each group is highlighted in \textbf{bold}.}
\label{tab:main_experiment}
\end{table*}

\subsubsection{Query-aware Selection}
To identify training samples that effectively improve reasoning ability, we first focus on queries whose predictions are significantly influenced by the presence of CoT. Intuitively, if a query leads to incorrect predictions without CoT but is answered correctly with CoT, it suggests that the query is non-trivial and that the CoT provides meaningful guidance. Such queries are more valuable for training, as they highlight reasoning challenges that benefit from exemplified thought processes.

Specifically, for each query $q_n$, we first evaluate the performance of the LLMPlayer without any CoT assistance. To ensure consistency with the CoT-aided setting, we perform $M \times K$ inference rounds using different sampling seeds:
\begin{equation}
\tilde{{v}}_{n, m, k} = \text{LLMPlayer}(q_n; \varepsilon_{n,m,k}).
\end{equation}

We then compare the LLMPlayer’s performance with and without CoT by computing the number of correct responses in each query. Specifically, for the CoT-free setting:
\begin{equation}
\tilde{\alpha}_n = \sum\nolimits_{m,k} \text{LLMJudge}(\tilde{{v}}_{n,m,k}, a_n).
\end{equation}

For the CoT-aided setting:
\begin{equation}
\alpha_n = \sum\nolimits_{m,k} \text{LLMJudge}({v}_{n,m,k}, a_n).
\end{equation}

The accuracy gain brought by CoT is defined as: $\Delta_{\alpha} = \alpha_n - \tilde{\alpha}_n$, which serves as a proxy for the training value of query $q_n$. A larger gain indicates that the query benefits more from reasoning guidance and is therefore more informative for improving the model’s reasoning capabilities.

However, the accuracy gain has a limitation: It does not fully capture the inherent difficulty of each query. For example, suppose $M = 3$ and $K = 6$, resulting in 18 CoTs per query. Consider two queries with the same accuracy gain $\alpha_n - \tilde{\alpha}_n$: $12-10$ and $5-3$. Although both exhibit a gain of $2$, the latter is clearly more challenging. Such query are arguably more valuable for fostering the model’s complex reasoning ability. To better reflect this aspect, we further introduce the CoT-aware selection.

\subsubsection{CoT-aware Selection}
\label{sec:cot-aware}
Given the generated dataset $\mathcal{D}_\text{cot}$, each query is associated with a best-performing CoT. If the model exhibits higher confidence when responding with it compared to without any CoT, this implies a stronger belief in its reasoning process, which is desirable for effective learning. Therefore, we define the following indicators based on Eq(\ref{confidence}):
\begin{equation}
    \begin{aligned}
        \beta_n^* &= \phi({v}_{n,\ m^*,k^*}), \quad \tilde{\beta}_{n,m,k} = \phi(\tilde{{v}}_{n,m,k}),\\
        \Delta_{\beta} &= \beta_n^* -\frac{1}{MK} \sum\nolimits_{m,k} \tilde{\beta}_{n,m,k},
    \end{aligned}
\end{equation}
where $\beta_n^*$ denotes the confidence of the model's best CoT-aided prediction, while the second term is the average confidence over all non-CoT predictions. The difference $\Delta_{\beta}$ captures the confidence gain provided by the CoT. A higher $\Delta_{\beta}$ indicates that the CoT significantly boosts the model’s belief in its answer, highlighting the CoT’s effectiveness in guiding the reasoning process.

Besides, we further introduce a correctness reward, defined as $f(v,a) = 2\cdot\text{LLMJudge}(v, a) - 1$. This is a constant term that takes the value of $+1$ when the model prediction $v$ matches the correct answer $a$, and $-1$ otherwise, since $\text{LLMJudge}(v, a)$ equals 1 for correct predictions and 0 for incorrect ones (i.e., Eq(\ref{LLMJudge})). Then, the final reward score is defined as:
\begin{equation}
    \begin{aligned}
        \gamma_{n}^* &= f({v}_{n,\ m^*,k^*}, a_n), \quad  \tilde{\gamma}_{n,m,k} = f(\tilde{{v}}_{n,m,k}, a_n), \\
        \Delta_{\gamma} &= \gamma_{n}^* -\frac{1}{MK} \sum\nolimits_{m,k} \tilde{\gamma}_{n,m,k}.
    \end{aligned}
\end{equation}

This score ensures that LLMPlayer receives higher rewards only when it correctly answers queries, thus making the selection process more meaningful.

To further illustrate the effect of the correctness reward, consider the same example used in Query-aware Selection, where two queries share the same accuracy gain. We first suppose the best CoT in both cases yields a correct answer, i.e., $\gamma_n^* = 1$. Then, for the first query where $\tilde{\alpha}_n = 10$ (10 correct, 8 incorrect among 18 CoTs), we have: $
\Delta_{\gamma} = 1 - \frac{1}{18}(10 - 8) = \frac{8}{9}
$. Similarly, for the second query with $\tilde{\alpha}_n = 3$ (3 correct, 15 incorrect), we get:
$
\Delta_{\gamma} = 1 - \frac{1}{18}(3 - 15) = \frac{5}{3}.
$
Although both queries have the same accuracy gain, the latter receives a higher reward due to its higher difficulty, making it more likely to be selected for training.

Finally, combining the query-aware and the CoT-aware selection, we define the final selection score as:
\begin{equation}
S(n) = \lambda_{\alpha} \Delta_{\alpha} + \lambda_{\beta}\Delta_{\beta} + \lambda_{\gamma}\Delta_{\gamma}.
\end{equation}

Using the above formula, we can compute a selection score for each instance in the dataset $\mathcal{D}_\text{cot}$. We then rank all $N$ instances by their scores and select the top $N'$ instances to construct the refined dataset $\mathcal{D}'_\text{cot}$.

%% file: sec/5_experiments.tex
\section{Experiments}

\begin{figure*}[ht!]
    \centering
    \includegraphics[width=1\linewidth]{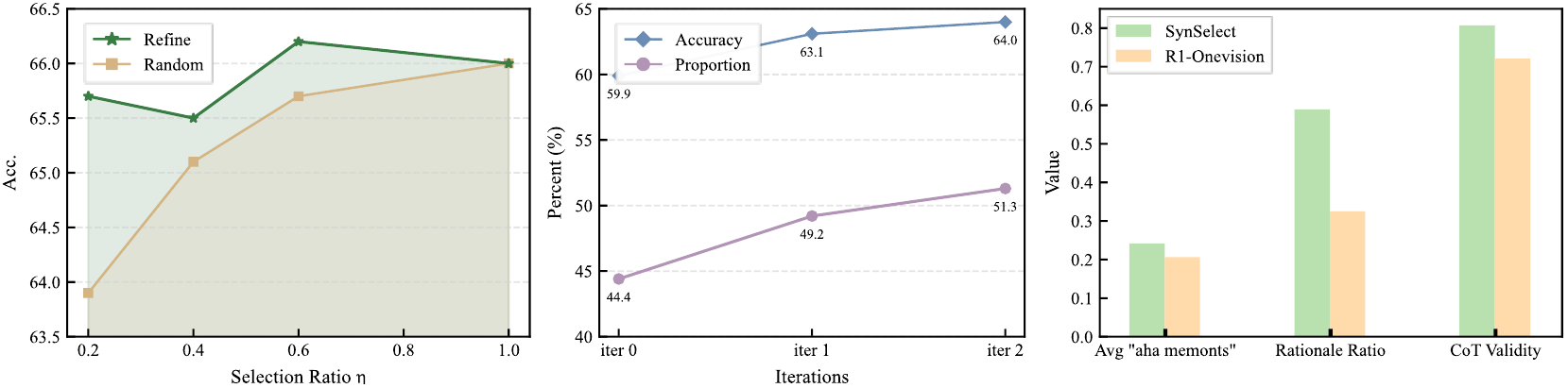}
    \caption{\textbf{(Left)} Sensitivity analysis of the selection ratio $\eta$. Experiments are conducted on the MathVista benchmark. \textbf{(Middle)} Results of the iterative bootstrap experiment. Evaluations are conducted on MathVista. \textbf{(Right)} Quantitative analysis of data quality. In addition to the Rationale Ratio defined in Section~\ref{sec:instance-level_select}, we also report the Avg ``aha moments'' and CoT Validity, as introduced in Section~\ref{sec:data_quality}.}
    \label{fig:combine}
\end{figure*}

\subsection{Experiment Settings}
\noindent
\textbf{Dataset and Benchmarks.} 
We conduct experiments on the R1-Onevision-Dataset \cite{yang2025r1}, a multimodal dataset containing 154,667 image-question-answer (IQA) instances, each paired with a CoT reasoning path that is synthesized with the vision-to-text conversion method. To construct our raw multimodal dataset $\mathcal{D}_{\text{raw}}$, we remove the original CoTs and retain only the IQA pairs.
After building $\mathcal{D}_\text{cot}$ and $\mathcal{D}'_\text{cot}$,  for fair comparison, we follow the same training pipeline as R1-Onevision, including SFT followed by RL optimization, to obtain MLRMs. Throughout all experiments, we use Qwen2.5-VL-7B \cite{bai_qwen25-vl_2025} as the model backbone. In the evaluation phase, we assess performance on several mainstream multimodal reasoning benchmarks, including MathVerse \cite{zhang2024mathverse}, MathVista \cite{lu2023mathvista}, WeMath \cite{qiao2024we}, and R1-Onevision-Bench \cite{yang2025r1}, using accuracy as the evaluation metric. Detailed descriptions of the datasets are provided in the Appendix.

\noindent
\textbf{Comparison Methods.} We compare our proposed method with the following models: (1) the baseline R1-Onevision model \cite{yang2025r1}; (2) the base model Qwen2.5-VL-7B\cite{bai_qwen25-vl_2025}; and (3) several representative state-of-the-art (SoTA) models, including GPT-4o \cite{hurst2024gpt}, InternVL2.5-8B \cite{chen2024expanding}, Qwen2-VL-7B \cite{Qwen2VL}, and LLaVA-CoT-11B \cite{xu_llava-cot_2025}. Among them, LLaVA-CoT-11B employs a four-stage CoT synthesis enhance its reasoning capabilities.

\noindent
\textbf{Implementation Details.} The Qwen2.5-VL series is employed as baseline models. For LLM Extractors, we use Qwen3-8B \cite{yang2025qwen3}. In experiments, we take $M=3$, $K=6$. The \textit{SynAgents} are R1-Onevision \cite{yang2025r1}, MM-Eureka \cite{meng2025mm}, and Vision-R1 \cite{huang2025vision}, respectively. For the score coefficients, we adopt $\lambda_k = 1, \lambda_\alpha = 2, \lambda_\beta = 1, \lambda_\gamma = 1$. 
During RL training (i.e., GRPO \cite{shao2024deepseekmath}), we employ CLEVR \cite{johnson2017clevr} as our training dataset.
Moreover, for evaluations, we conducted five independent runs.
Further details regarding dataset construction, training procedures, and benchmark specifications are available in the Appendix.

\subsection{Main Results}

\begin{figure}
    \centering
    \includegraphics[width=1\linewidth]{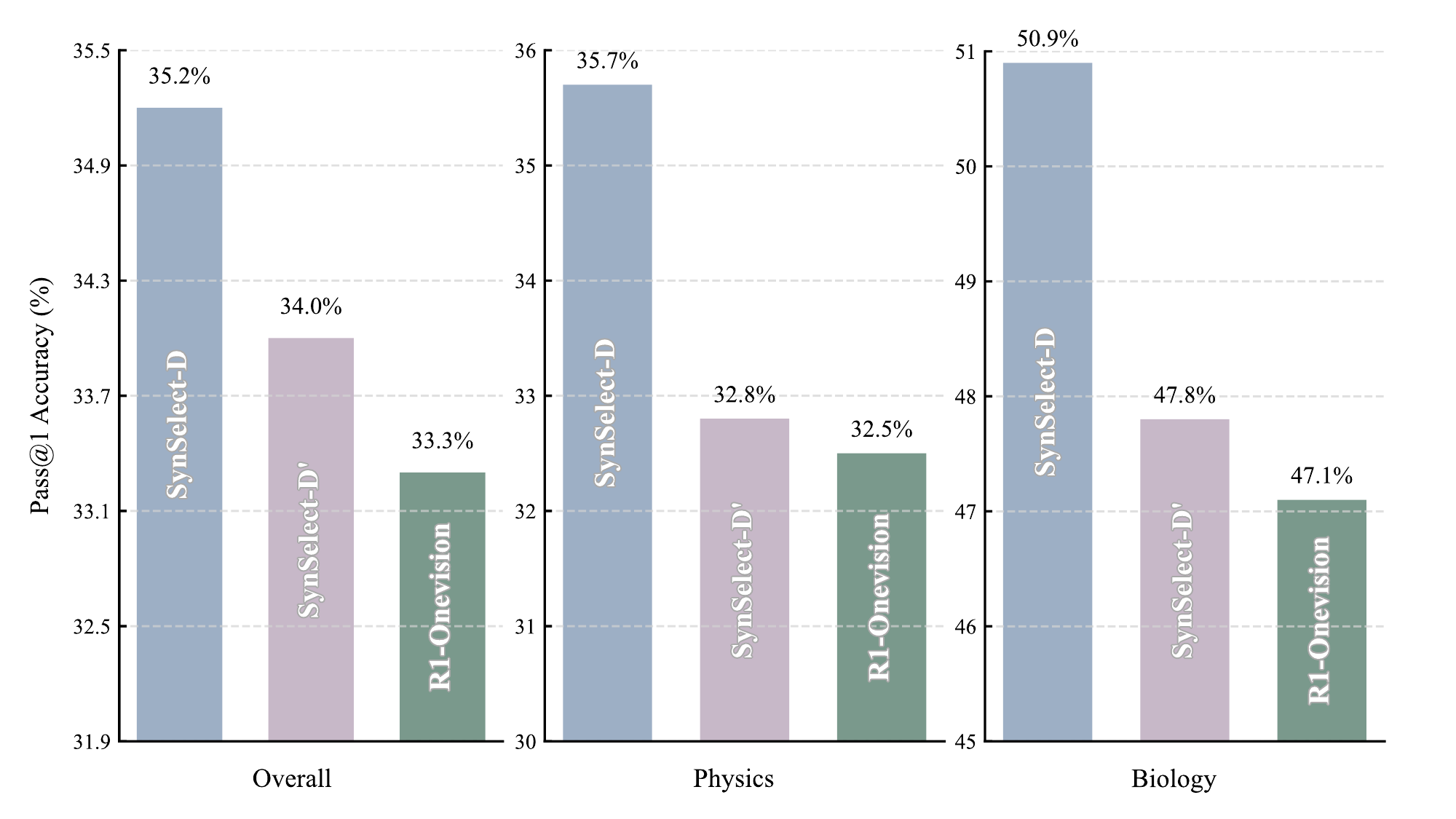}
    \caption{Experimental results on R1-Onevision-Bench.}
    \label{fig:r1ovbench}
\end{figure}



\noindent
\textbf{Model Performance.}
We first fine-tune MLLMs with CoT data synthesized by SynSelect, followed by RL optimization (GRPO).
As shown in Table \ref{tab:main_experiment}, models trained on both $\mathcal{D}_{\text{cot}}$ and $\mathcal{D}'_{\text{cot}}$  consistently outperform existing SoTA methods across most multimodal reasoning benchmarks.
Notably, SynSelect-$\mathcal{D}_{\text{cot}}$ achieves around 2\% gains over R1-Onevision on MathVista and WeMath, and even surpasses LLaVA-CoT-11B, demonstrating that the proposed CoT synthesis strategy is more effective than existing CoT construction methods for enhancing multimodal reasoning performance.
Moreover, $\mathcal{D}'_{\text{cot}}$ further improves performance, showing that a compact subset of high-quality CoTs can better enhance complex reasoning ability and training efficiency.

We further evaluate SynSelect on R1-Onevision-Bench, a multimodal reasoning benchmarkcovering a variety of domains such as physics, biology, and related areas. All models are trained under the same SFT+RL setup for fair comparison. As shown in Figure~\ref{fig:r1ovbench}, models trained on SynSelect-synthesized data consistently outperform those trained on the original R1-Onevision dataset across all domains, confirming that SynSelect-generated CoTs generalize effectively to diverse reasoning scenarios. Interestingly, unlike our earlier results on mathematical reasoning benchmarks, the full dataset $\mathcal{D}_{\text{cot}}$ yields the best performance on R1-Onevision-Bench. A plausible explanation is that larger-scale SFT data provides a wider exploration space, enabling the model to handle heterogeneous problem types more effectively.

\noindent
\textbf{Data Quality.}
\label{sec:data_quality}
Beyond performance, we assess the synthesized data from multiple perspectives, including “aha moments”, rationale ratio, and CoT validity.
As shown in Figure~\ref{fig:combine} (right), Avg ``aha moments'' indicates the average occurrence frequency of trigger tokens (i.e., ``Wait...'') in the data. Rationale ratio has been presented in Section \ref{sec:instance-level_select}. CoT Validity refers to the accuracy of LLMPlayer in answering questions when the corresponding CoT is provided as input.
Compared with R1-Onevision, SynSelect yields clear improvements across all metrics, especially in rationale ratio, indicating that it constructs more coherent and informative reasoning chains.

\noindent
\textbf{Analysis of Selection Mechanism.}
In this section, we investigate the effect of the selection ratio $\eta$ by setting $\eta = \{0.2, 0.4, 0.6, 1.0\}$. As shown in Figure~\ref{fig:combine} (left), models trained on the refined subset $\mathcal{D}'_{\text{cot}}$ consistently outperform randomly sampled data. Notably, when the sampling rate is 20\%, the performance of randomly sampled data drops significantly. Moreover, 20\% of selected data matches or exceeds the performance of 60\% of random data. These results clearly demonstrate that the quality of SFT data is critical for improving a model’s reasoning ability, and that an effective selection mechanism is essential for maximizing training efficiency and performance.

\subsection{Ablation Study}

In addition to evaluating models trained with SFT+RL, we also assess the performance of models trained with SFT alone. As shown in Table~\ref{tab:main_experiment}, models trained on \textit{SynSelect}-synthesized data without RL consistently outperform those trained on the original R1-Onevision dataset. On both MathVista and WeMath, our method achieves approximately a 4\% improvement over the baseline, further demonstrating the effectiveness of the proposed CoT synthesis approach.

\begin{table}[!h]
\centering

\scriptsize

\begin{tabularx}{\linewidth}{
  >{\raggedright\arraybackslash}p{1.9cm}  
  >{\centering\arraybackslash}X    
  >{\centering\arraybackslash}X    
  >{\centering\arraybackslash}X    
}
\toprule
\textbf{Model} & \textbf{MathVerse ALL} & \textbf{MathVerse Vision-only} & \textbf{MathVista} \\
\midrule
\multicolumn{4}{l}{\textit{\textbf{Base Model}}} \\  
Qwen2.5-VL  & 43.6 & 38.2 & 63.7 \\
\midrule
\multicolumn{4}{l}{\textit{\textbf{SFT Only}}} \\  
Qwen2.5-VL  & 43.1$\pm$1.0 & 38.5$\pm$0.9 & 64.5$\pm$1.5 \\
\textit{SynSelect}-$\mathcal{D}_\text{cot}$ & \textbf{43.7$\pm$0.2} & 39.1$\pm$0.7 & \textbf{65.6$\pm$1.6} \\
\textit{SynSelect}-$\mathcal{D}'_\text{cot}$ & 43.4$\pm$0.3 & \textbf{39.5$\pm$0.9} & 64.7$\pm$1.4 \\
\midrule
\multicolumn{4}{l}{\textit{\textbf{SFT + RL}}} \\  
Qwen2.5-VL & 44.2$\pm$1.3 & 39.7$\pm$1.6 & 66.8$\pm$1.8 \\
\textit{SynSelect}-$\mathcal{D}_\text{cot}$ & 45.7$\pm$2.2 & 40.9$\pm$2.1 & 66.3$\pm$1.0 \\
\textit{SynSelect}-$\mathcal{D}'_\text{cot}$ & \textbf{46.9$\pm$1.5} & \textbf{42.4$\pm$2.4} & \textbf{67.1$\pm$2.6} \\
\bottomrule
\end{tabularx}
\caption{Experimental results on multimodal reasoning benchmarks. Unlike Table~\ref{tab:main_experiment}, we have changed the raw dataset and taken Vision-R1-Cold as $\mathcal{D}_\text{raw}$. Results are reported as (mean$\pm$std). The highest accuracy in each group is highlighted in \textbf{bold}.}
\label{tab:vrc_experiment}
\end{table}


\subsection{Scalability}

To assess the scalability of \textit{SynSelect}, we further replace the original raw dataset $\mathcal{D}_\text{raw}$ with Vision-R1-Cold~\cite{huang2025vision}, which differs from R1-OneVision-Dataset in data distribution and modality conversion strategy. This experiment examines whether \textit{SynSelect} generalizes to new data sources. As shown in Table~\ref{tab:vrc_experiment}, \textit{SynSelect} consistently improves performance across all benchmarks under this new setting. Both \textit{SynSelect}-$\mathcal{D}_{\text{cot}}$ and \textit{SynSelect}-$\mathcal{D}_{\text{cot}}^{\prime}$ outperform the Qwen2.5-VL baseline in SFT-only and SFT+RL stages, achieving the highest accuracies on MathVerse and MathVista. These results confirm that SynSelect’s synthesis–selection mechanism generalizes well across datasets, demonstrating strong scalability and robustness in multimodal reasoning tasks.


\subsection{Iterative Bootstrap}

We further explore whether SynSelect can self-improve through iterative refinement. Starting from a model fine-tuned on R1-OneVision-Dataset, we repeatedly use it as SynAgent to synthesize new CoT data, fine-tune the base model, and replace the SynAgent in a closed loop of ``generate $\rightarrow$ fine-tune $\rightarrow$ regenerate.''

As shown in Figure~\ref{fig:combine} (middle), both reasoning accuracy and the proportion of self-generated data increase consistently across iterations—from 59.9\% to 64.0\% and 44.4\% to 51.3\%, respectively. This trend reveals a positive feedback loop between model improvement and data quality, demonstrating that SynSelect acts as a self-bootstrapping data engine capable of continual autonomous enhancement.

%% file: sec/6_conclusion.tex
\section{Conclusion}

In this paper, we proposed \textit{SynSelect}, a three-stage synthesis-selection framework for constructing high-quality long CoT data tailored to multimodal reasoning tasks. Specifically, \textit{SynSelect} first synthesized comprehensive and diverse reasoning paths by leveraging multiple heterogeneous MLRMs. Then, we employed an instance-level selection strategy to identify the most suitable CoT for each instance based on answer correctness, reasoning validity, and length appropriateness metrics. Finally, we selected a compact and instructive subset through batch selection strategies, including query-aware and CoT-aware selection methods.
Experimental results on multiple multimodal benchmarks demonstrated the effectiveness of our framework.

%% file: sec/X_supp.tex
\clearpage
\setcounter{page}{1}
\maketitlesupplementary

\section{Details of Dataset and Benchmarks}
Our framework is built upon the R1-Onevision-Dataset~\cite{yang2025r1}, and the trained models are evaluated across multiple multimodal reasoning benchmarks, including MathVista~\cite{lu2023mathvista}, MathVerse~\cite{zhang2024mathverse}, and WeMath~\cite{qiao2024we}:

\paragraph{Dataset} R1-Onevision-Dataset~\cite{yang2025r1} is a multimodal reasoning dataset that includes samples from domains such as science, mathematics, charts, and natural scenarios, with a total of over 155k examples. Each sample contains an image-question-answer (IQA) pair along with a Chain-of-Thought (CoT). The original CoTs were generated using a text-based reasoning model (e.g., DeepSeek-R1~\cite{deepseek-ai_deepseek-r1_2025}) through an image-to-text conversion method. In this work, we remove the original CoTs and regenerate them using the \textit{SynSelect} framework.

\paragraph{Benchmarks}
\begin{itemize}
    \item MathVista~\cite{lu2023mathvista}: a mathematical benchmark constructed to integrate challenges across diverse mathematical and visual tasks. Its Test Mini split, containing approximately 1,000 samples, is utilized in our evaluation. 
    \item MathVerse~\cite{zhang2024mathverse}: a comprehensive visual mathematics benchmark developed for equitable and in-depth assessment of multimodal large language models (MLLMs). We utilize the full dataset and additionally report the ``Vision-Only'' results, which expose substantial challenges by incorporating the entire question within the diagram. 
    \item WeMath~\cite{qiao2024we}: a benchmark designed to explore problem-solving mechanisms beyond end-to-end performance. We adopt its Test Mini split, comprising around 1,740 samples, with average accuracy serving as our primary reporting metric. 
    \item R1-Onevision-Bench~\cite{yang2025r1}: a comprehensive multimodal reasoning benchmark covering mathematics, physics, chemistry, biology, and logical deduction across 38 subcategories. It comprises 942 problems paired with images. We adopt the full benchmark and report average accuracy as the primary metric. 
\end{itemize}

\section{Data Analysis}

\begin{figure}[!t]
    \centering
    \includegraphics[width=1\linewidth]{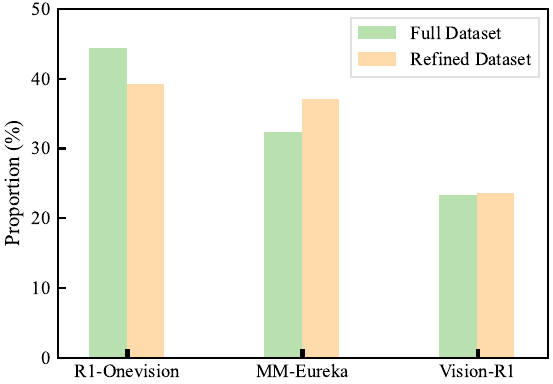}
    \caption{Source Distribution of Synthesized CoTs}
    \label{fig:portion}
\end{figure}

In this section, we present the basic statistics of the SynSelect dataset. 

The full synthesized dataset $\mathcal{D}_\text{cot}$ consists of 150,435 CoTs, with an average length of 226.4 tokens and a standard deviation of 176.9. Among them, 66,864 CoTs are generated by R1-Onevision~\cite{yang2025r1}, while MM-Eureka~\cite{meng2025mm} and Vision-R1~\cite{huang2025vision} contribute 48,522 and 35,044 samples, respectively.

The refined dataset $\mathcal{D}'_\text{cot}$ contains 30,087 CoTs, with an average length of 196.5 tokens and a standard deviation of 133.5. Within this subset, 11,838 CoTs are from R1-Onevision~\cite{yang2025r1}, while MM-Eureka~\cite{meng2025mm} and Vision-R1~\cite{huang2025vision} contribute 11,175 and 7,074 CoTs, respectively.

Figure~\ref{fig:portion} illustrates the proportion of CoTs contributed by each SynAgent in both datasets. In the full dataset $\mathcal{D}_\text{cot}$, R1-Onevision accounts for a relatively larger share. However, after refinement, the contributions become more balanced across different models. This shift suggests that the \textit{SynSelect} framework is effective in leveraging the strengths of multiple models, resulting in a higher-quality and more diverse reasoning dataset.




\section{Illustration of the \textit{SynSelect} Pipeline}

\begin{figure*}[!ht]
    \centering
    \includegraphics[width=0.9\linewidth]{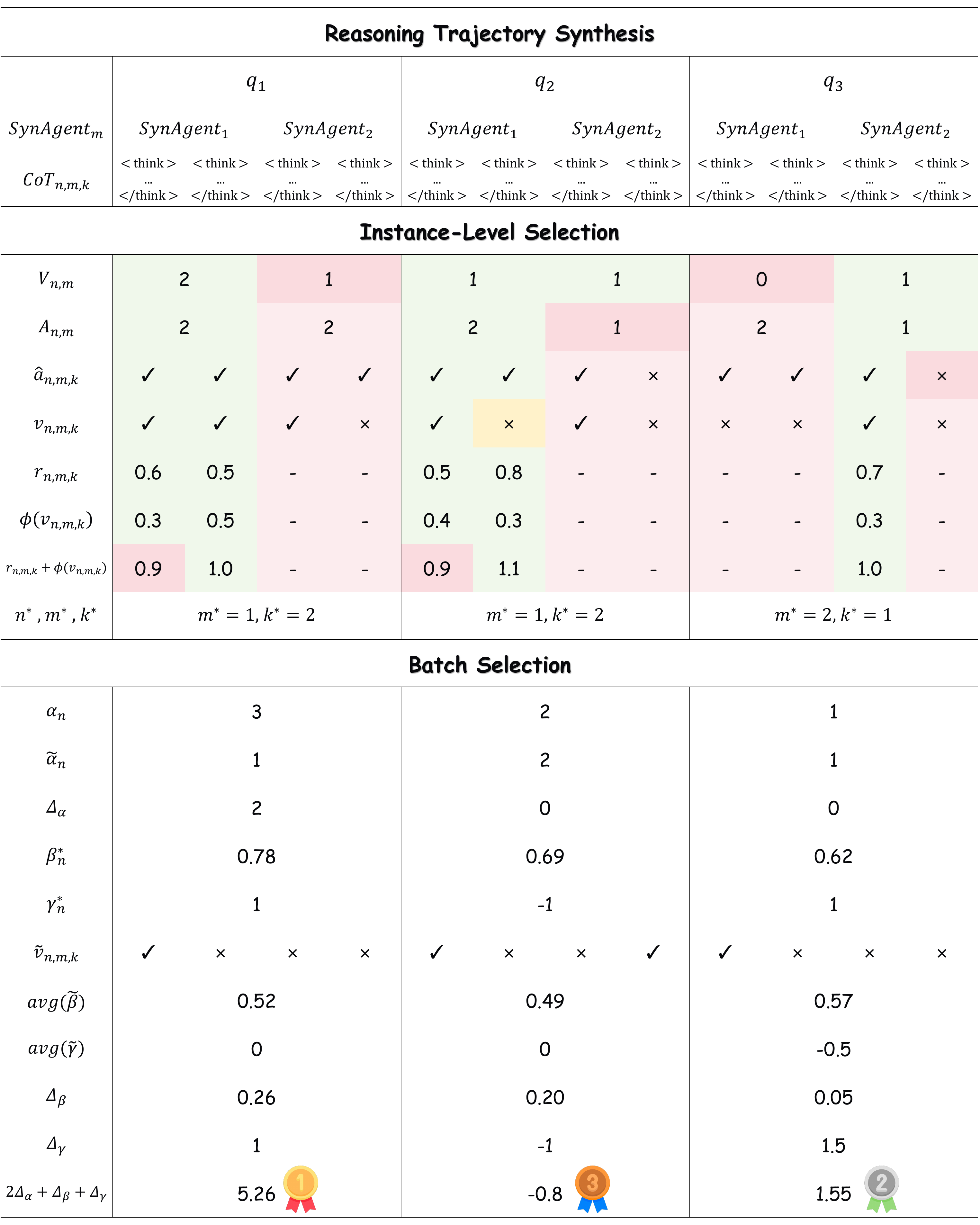}
    \caption{Illustration of the \textit{SynSelect} pipeline. We use a pink background to represent the branches that have been eliminated.}
    \label{fig:pipeline}
\end{figure*}

To illustrate the computation process of the \textit{SynSelect} framework, we provide a simplified example. For ease of explanation, in Figure~\ref{fig:pipeline}, we assume that two SynAgents are used, each generating two candidate CoTs, i.e., $M = 2$ and $K = 2$.

As shown in Figure~\ref{fig:pipeline}, in the \textit{\textbf{Reason Trajectory Synthesis}} phase, we generate $M \times K = 2 \times 2 = 4$ candidate CoTs for each query.

Next, during the \textit{\textbf{Instance-Level Selection}} phase, we select the optimal CoT for each query, resulting in the synthesized reasoning dataset $\mathcal{D}_\text{cot}$.

Finally, in the \textit{\textbf{Batch Selection}} phase, we rank all queries based on the weighted score $\lambda_\alpha\Delta_\alpha + \lambda_\beta\Delta_\beta + \lambda_\gamma\Delta_\gamma$, and select the top-$N'$ instances according to the selection ratio $\eta$, yielding the refined reasoning dataset $\mathcal{D'}_\text{cot}$.

\section{Case Study}

\subsection{\textit{SynSelect} Data}

\begin{figure*}
    \centering
    \includegraphics[width=\linewidth]{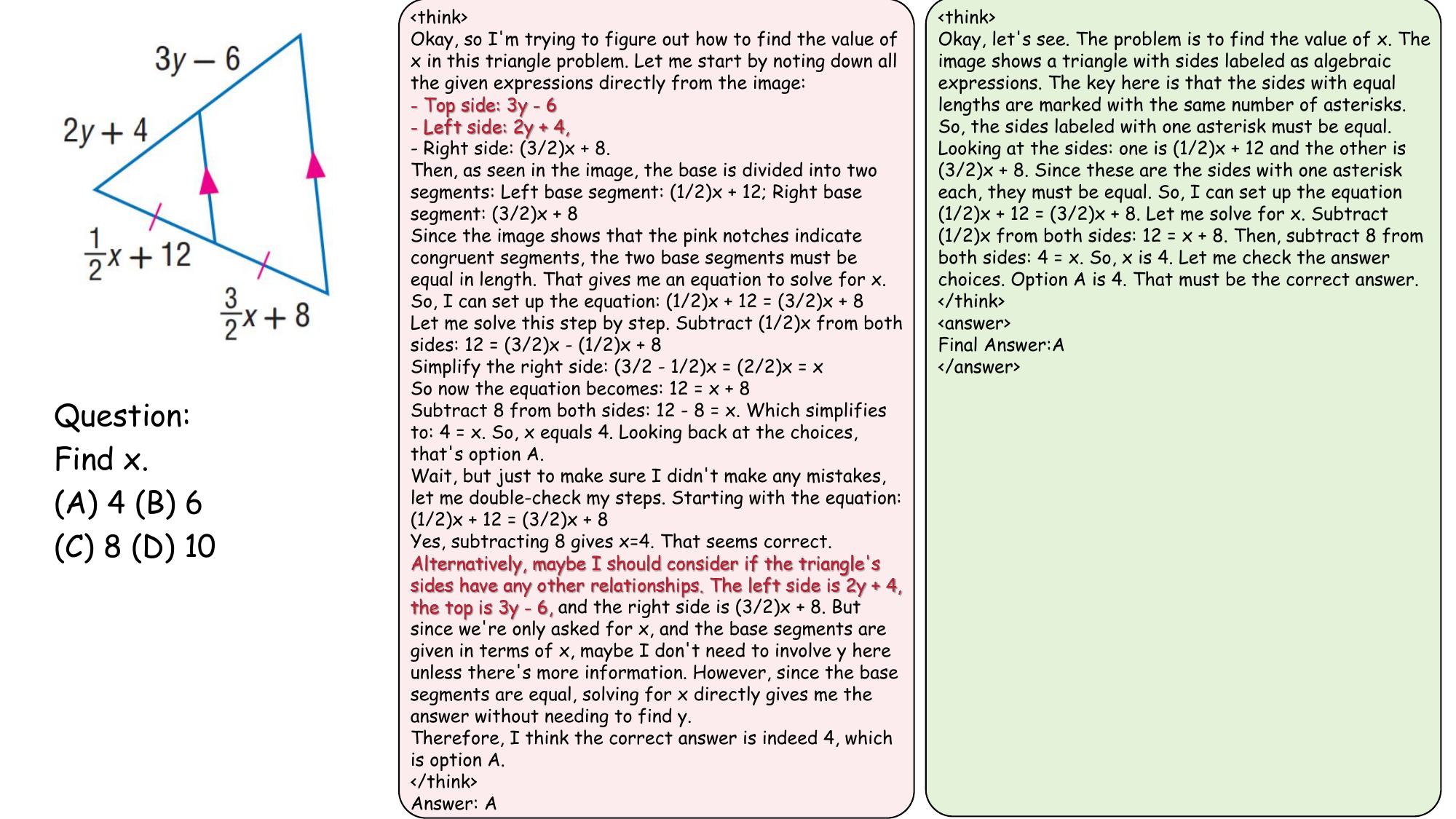}
    \caption{Case study 1 of \textit{SynSelect} Data. Left is the R1-Onevision data and right is \textit{SynSelect} Data. Red boldface indicates redundant reasoning processes.}
    \label{fig:case_study_1-1}
\end{figure*}

\begin{figure*}
    \centering
    \includegraphics[width=1\linewidth]{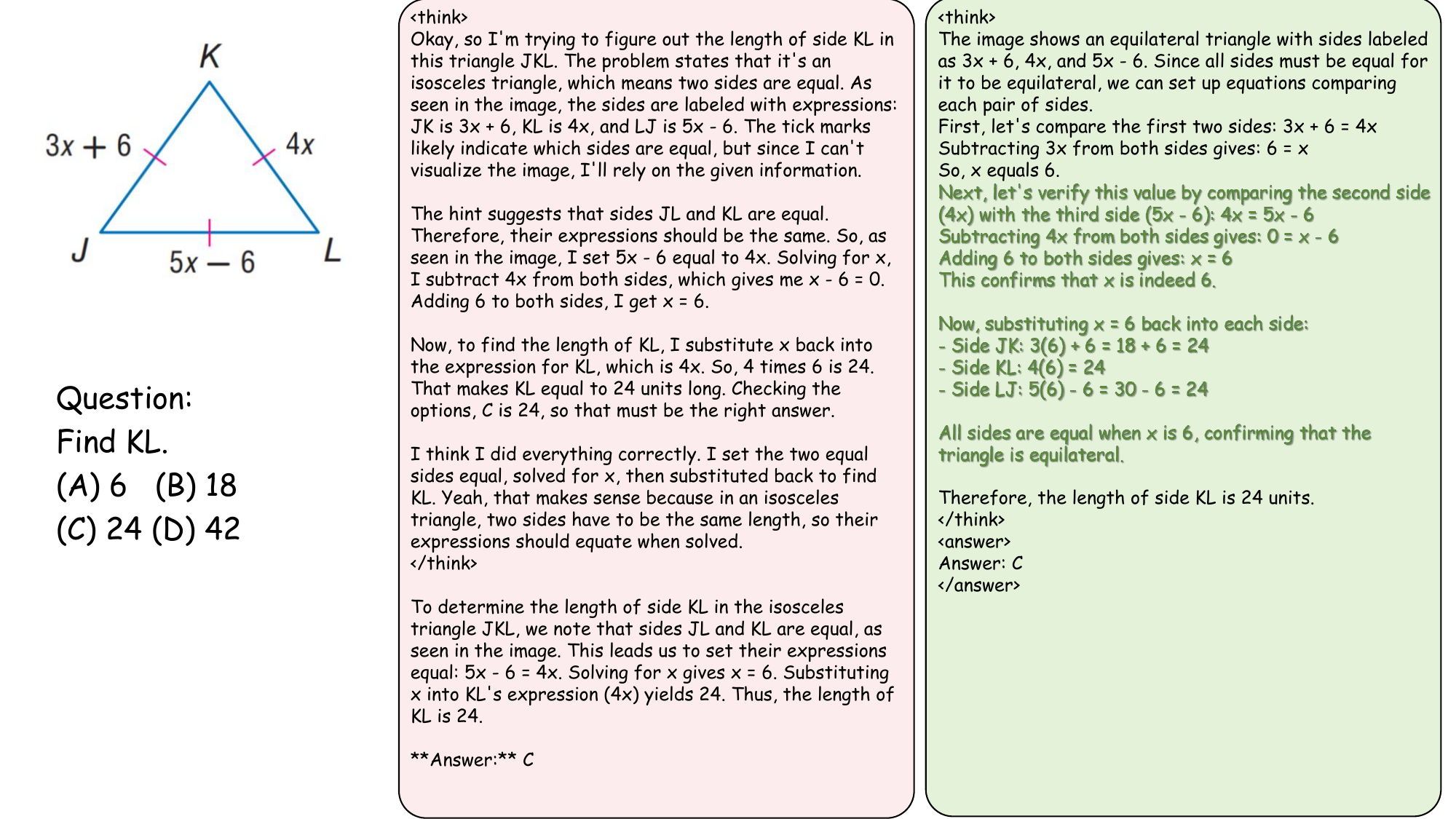}
    \caption{Case study 2 of \textit{SynSelect} Data. Left is the R1-Onevision data and right is \textit{SynSelect} Data. Green boldface indicates effective inspection processes.}
    \label{fig:case_study_1-2}
\end{figure*}

\begin{figure*}
    \centering
    \includegraphics[width=1\linewidth]{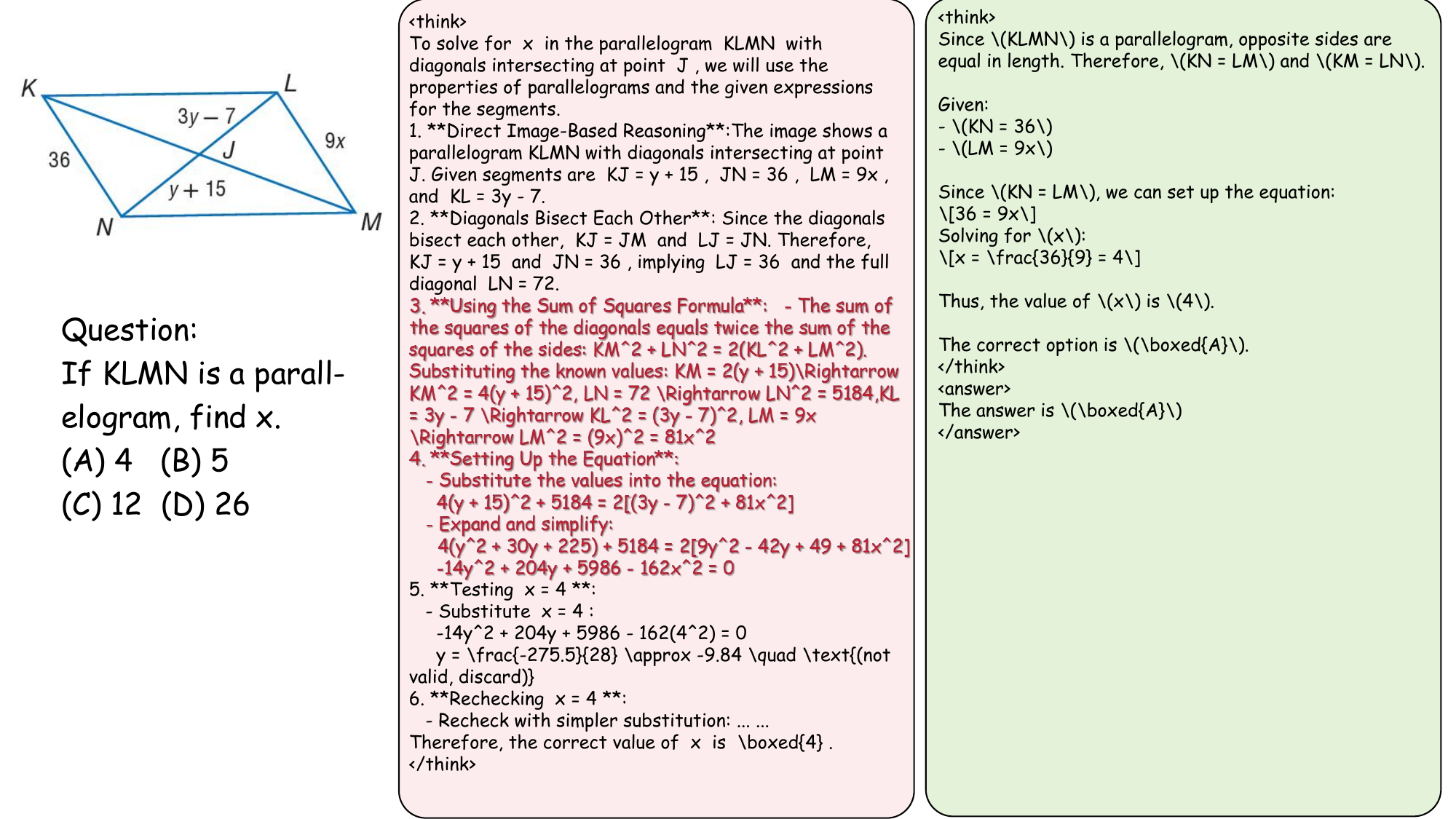}
    \caption{Case study 3 of \textit{SynSelect} Data. Left is the R1-Onevision data and right is \textit{SynSelect} Data. Red boldface indicates redundant reasoning processes.}
    \label{fig:case_study_1-3}
\end{figure*}

\begin{figure*}
    \centering
    \includegraphics[width=1\linewidth]{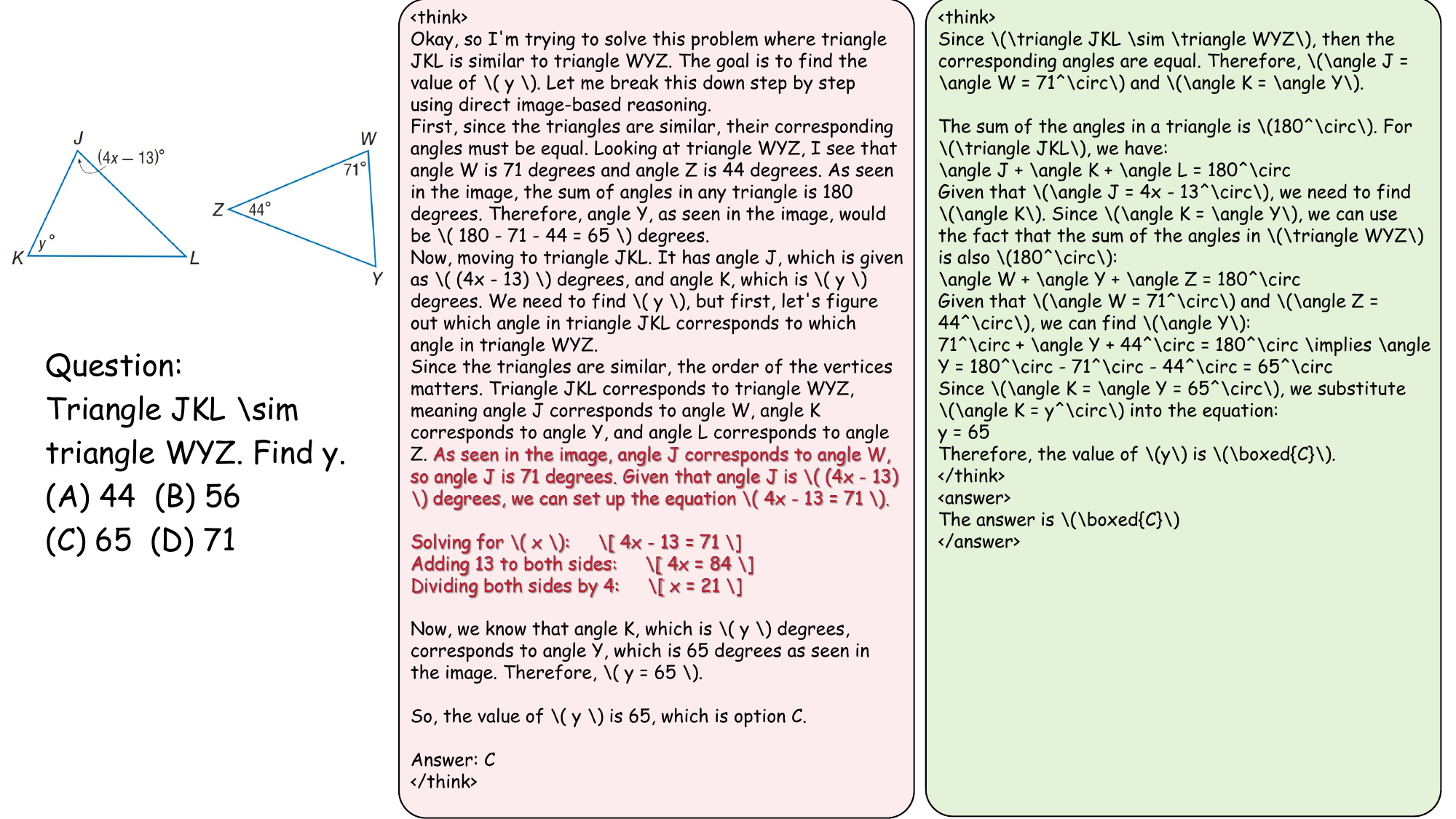}
    \caption{Case study 4 of \textit{SynSelect} Data. Left is the R1-Onevision data and right is \textit{SynSelect} Data. Red boldface indicates redundant reasoning processes.}
    \label{fig:case_study_1-4}
\end{figure*}

\begin{figure*}
    \centering
    \includegraphics[width=1\linewidth]{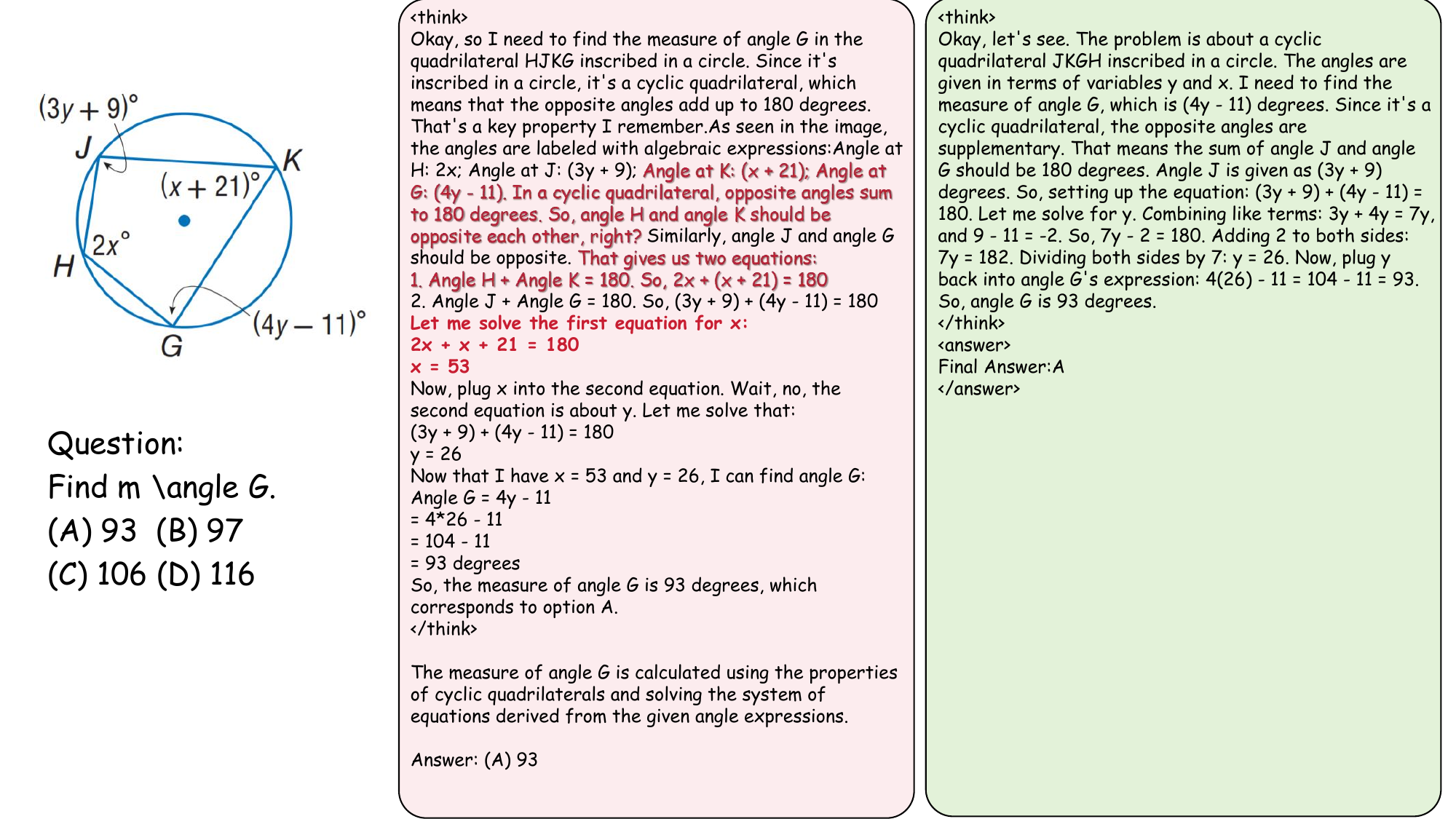}
    \caption{Case study 5 of \textit{SynSelect} Data. Left is the R1-Onevision data and right is \textit{SynSelect} Data. Red boldface indicates redundant reasoning processes.}
    \label{fig:case_study_1-5}
\end{figure*}

In this section, we present a case study on \textit{SynSelect}-synthesized Data and compare them with the original CoTs in R1-Onevision-Dataset. Specific examples are provided in Figure~\ref{fig:case_study_1-1}-\ref{fig:case_study_1-5}.

\subsection{Model Outputs}

\begin{figure*}
    \centering
    \includegraphics[width=1\linewidth]{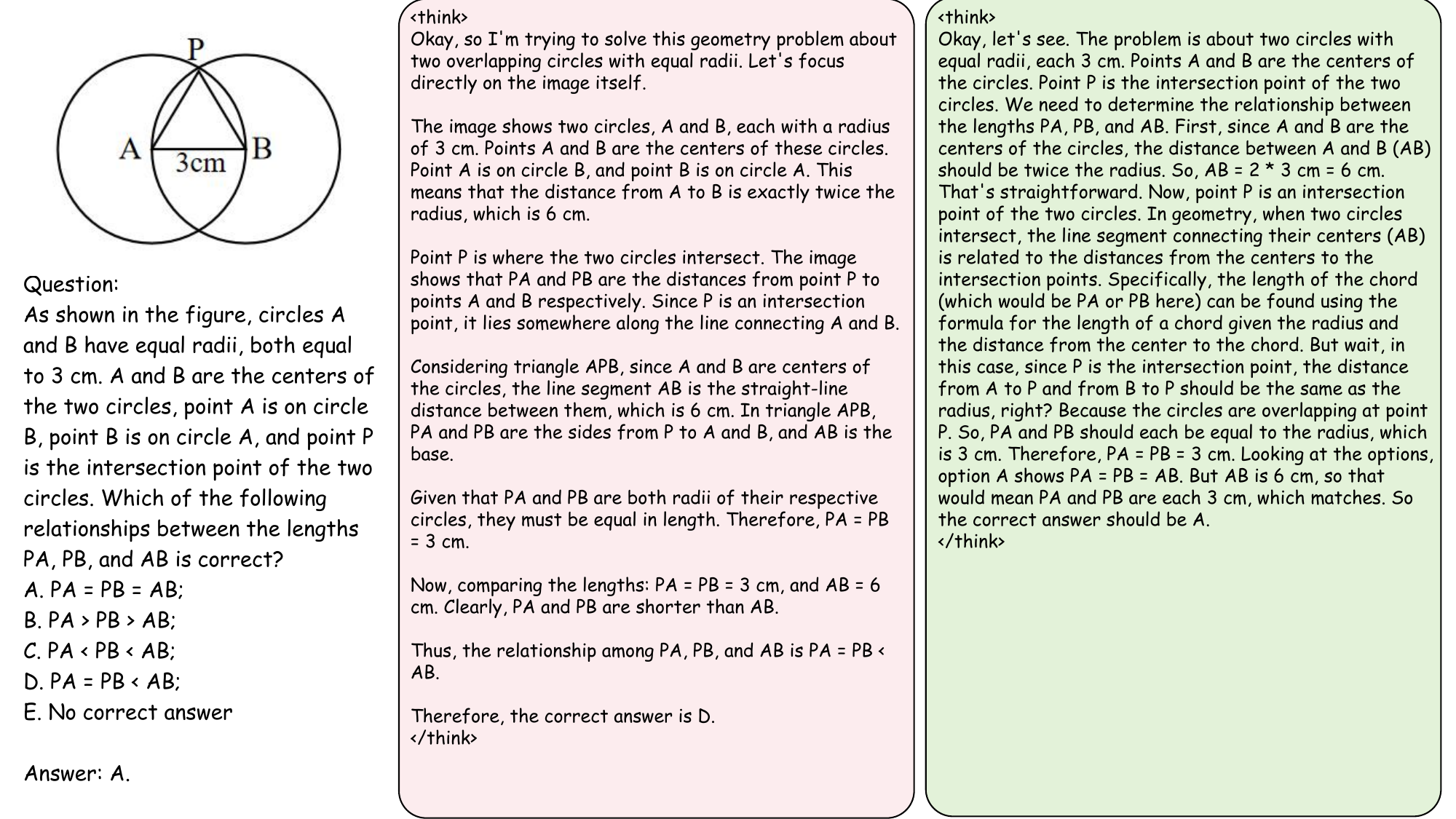}
    \caption{Case study 1 of Model Outputs. Left is the baseline (R1-Onevision-7B-RL) and right is the model which is trained on \textit{SynSelect}-Synthesized data.}
    \label{fig:case_study_2-1}
\end{figure*}

\begin{figure*}
    \centering
    \includegraphics[width=1\linewidth]{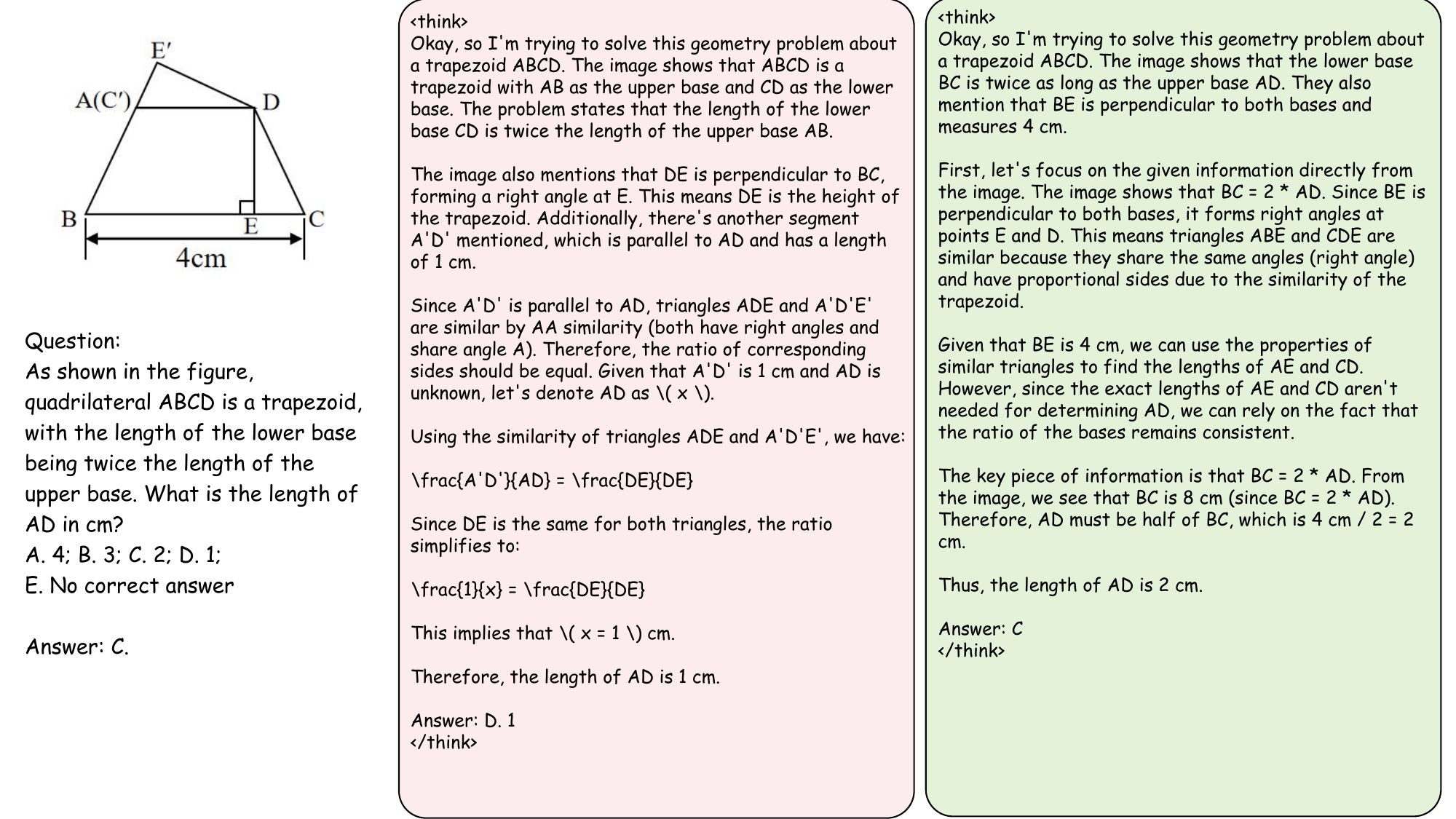}
    \caption{Case study 2 of Model Outputs. Left is the baseline (R1-Onevision-7B-RL) and right is the model which is trained on \textit{SynSelect}-Synthesized data.}
    \label{fig:case_study_2-2}
\end{figure*}

\begin{figure*}
    \centering
    \includegraphics[width=1\linewidth]{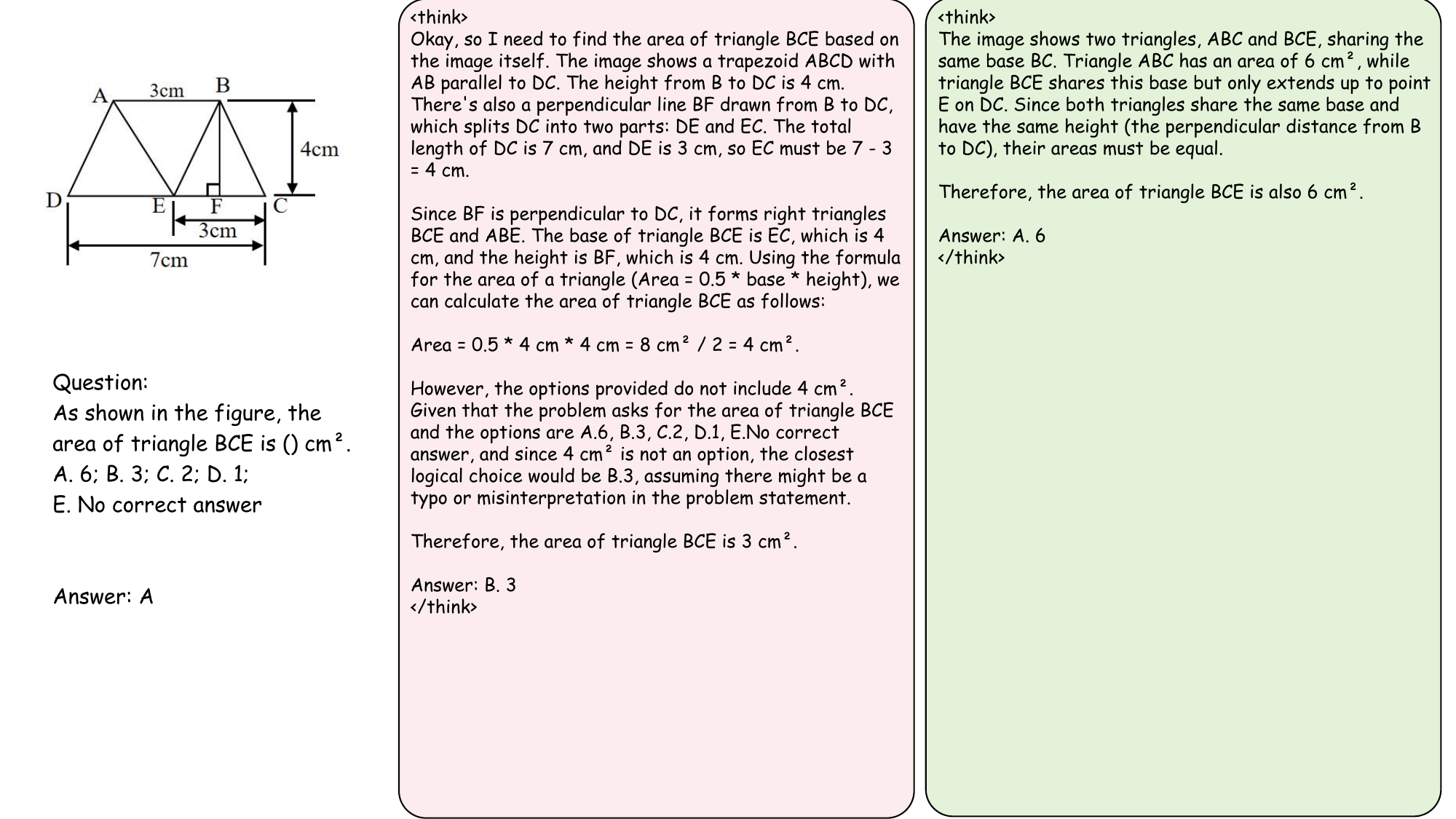}
    \caption{Case study 3 of Model Outputs. Left is the baseline (R1-Onevision-7B-RL) and right is the model which is trained on \textit{SynSelect}-Synthesized data.}
    \label{fig:case_study_2-3}
\end{figure*}

\begin{figure*}
    \centering
    \includegraphics[width=1\linewidth]{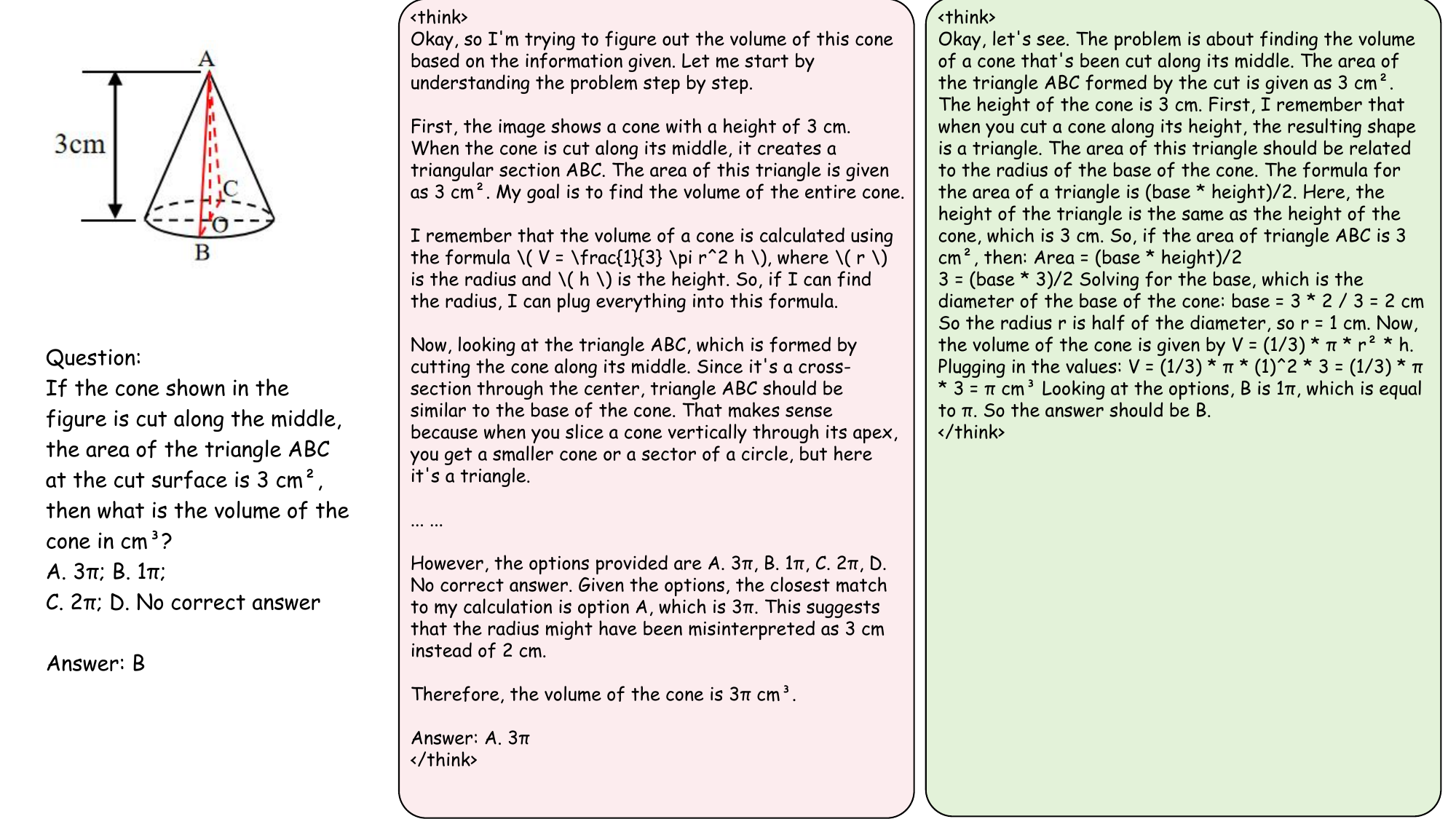}
    \caption{Case study 4 of Model Outputs. Left is the baseline (R1-Onevision-7B-RL) and right is the model which is trained on \textit{SynSelect}-Synthesized data.}
    \label{fig:case_study_2-4}
\end{figure*}

\begin{figure*}
    \centering
    \includegraphics[width=1\linewidth]{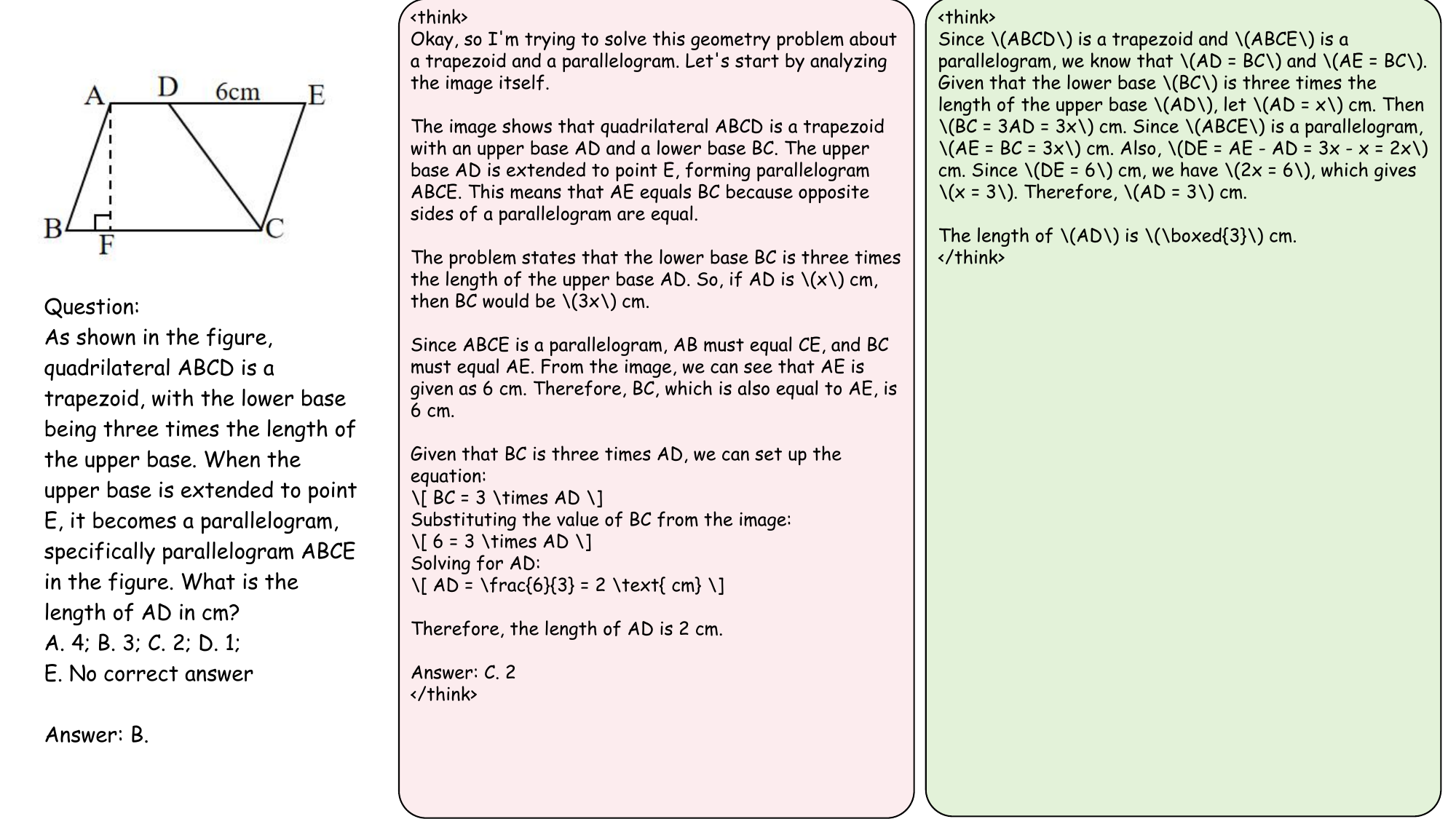}
    \caption{Case study 5 of Model Outputs. Left is the baseline (R1-Onevision-7B-RL) and right is the model which is trained on \textit{SynSelect}-Synthesized data.}
    \label{fig:case_study_2-5}
\end{figure*}

In this section, we present a case study on the performance of the model trained on \textit{SynSelect}-synthesized data (SFT+RL), and compare it with the baseline R1-Onevision-7B-RL~\cite{yang2025r1}. Specific examples are provided in Figure~\ref{fig:case_study_2-1}-\ref{fig:case_study_2-5}.

